\documentclass{article}

\usepackage{microtype}
\usepackage{graphicx}
\usepackage{subfigure}
\usepackage{booktabs} 

\usepackage{hyperref}

\usepackage{threeparttable}
\usepackage{enumitem}
\usepackage{amsmath}

\usepackage[accepted]{icml2019}

\icmltitlerunning{IPC: A Benchmark Data Set for Learning with Graph-Structured Data}

\begin{document}

\twocolumn[
\icmltitle{IPC: A Benchmark Data Set for Learning with Graph-Structured Data}



\icmlsetsymbol{equal}{*}

\begin{icmlauthorlist}
\icmlauthor{Patrick Ferber}{basel}
\icmlauthor{Tengfei Ma}{ibm}
\icmlauthor{Siyu Huo}{ibm}
\icmlauthor{Jie Chen}{ibm,mitibm}
\icmlauthor{Michael Katz}{ibm}
\end{icmlauthorlist}

\icmlaffiliation{basel}{University of Basel}
\icmlaffiliation{ibm}{IBM Research}
\icmlaffiliation{mitibm}{MIT-IBM Watson AI Lab}

\icmlcorrespondingauthor{Jie Chen}{chenjie@us.ibm.com}
\icmlcorrespondingauthor{Michael Katz}{Michael.Katz1@ibm.com}

\icmlkeywords{IPC, graph data set}

\vskip 0.3in
]



\printAffiliationsAndNotice{}  

\begin{abstract}
Benchmark data sets are an indispensable ingredient of the evaluation of graph-based machine learning methods. We release a new data set, compiled from International Planning Competitions (IPC), for benchmarking graph classification, regression, and related tasks. Apart from the graph construction (based on AI planning problems) that is interesting in its own right, the data set possesses distinctly different characteristics from popularly used benchmarks. The data set, named IPC, consists of two self-contained versions, grounded and lifted, both including graphs of large and skewedly distributed sizes, posing substantial challenges for the computation of graph models such as graph kernels and graph neural networks. The graphs in this data set are directed and the lifted version is acyclic, offering the opportunity of benchmarking specialized models for directed (acyclic) structures. Moreover, the graph generator and the labeling are computer programmed; thus, the data set may be extended easily if a larger scale is desired. The data set is accessible from \url{https://github.com/IBM/IPC-graph-data}.
\end{abstract}

\section{Introduction}
Benchmark data sets are indispensable in the evaluation of machine learning
models for graph-structured data. With the surging interest in graph
representation learning, a rich collection of data sets constructed from
real-life applications becomes important for the validation of effectiveness of
any existing or newly proposed method, and for demonstrating its widespread
applicability. We introduce a new labeled data set, IPC, compiled from AI plannig tasks described in the
Planning Domain Definition Language (PDDL) \citep{mcdermott-aimag2000}. 

In this data set, each planning task is represented as a directed graph, which
has target values
and whose nodes
are equipped with features. Planning tasks described in PDDL
admit a concise representation in transition graphs, however too big to fit in
any conceivable size memory. 
Recent advances in planning allow to encode some structural information of a
 task in graphs of manageable size. Two examples are the problem
description graph~\citep{pochter-et-al-aaai2011}, for a \emph{grounded} task
representation, and the abstract structure
graph~\citep{sievers-et-al-icaps2019}, for a \emph{lifted} representation.
Hence, our data set consists of two versions of graphs (IPC-grounded and
IPC-lifted) for the same set of tasks; each version may be used independently.
There are 2439 planning tasks in total, pre-split for training, validation, and
testing.
Moreover, the lifted version is acyclic.

Accompanied with the tasks are performance results for $17$ cost-optimal
domain-independent planners, each of which attempts to solve a task under a
timeout limit $T=1800$ seconds. Hence, the target values for each graph are the
CPU times of these planners, gathered on the same hardware. For practical
reasons, if a planner cannot solve the task before timeout, the target value is
artificially set as $10000$.

The background on AI Planning and graph construction, including example problem
domains, node feature definition, and how the data set can be extended, are
presented in Section~\ref{sec:construction}.
The characteristics of this data set are different in several aspects from those
of the commonly used benchmark data sets for graph kernels and graph neural
networks: The sizes of our graphs not only are substantially larger but also
vary significantly. The imposed challenges and implications
are elaborated in Section~\ref{sec:stat}.
We present an example use of the data set in Section~\ref{sec:use} and conclude
in Section~\ref{sec:conclude}.

\section{Data Set Construction}\label{sec:construction}

\begin{figure*}[ht]
\centering
\subfigure[Problem (in PDDL)]{\includegraphics[height=1.3in]{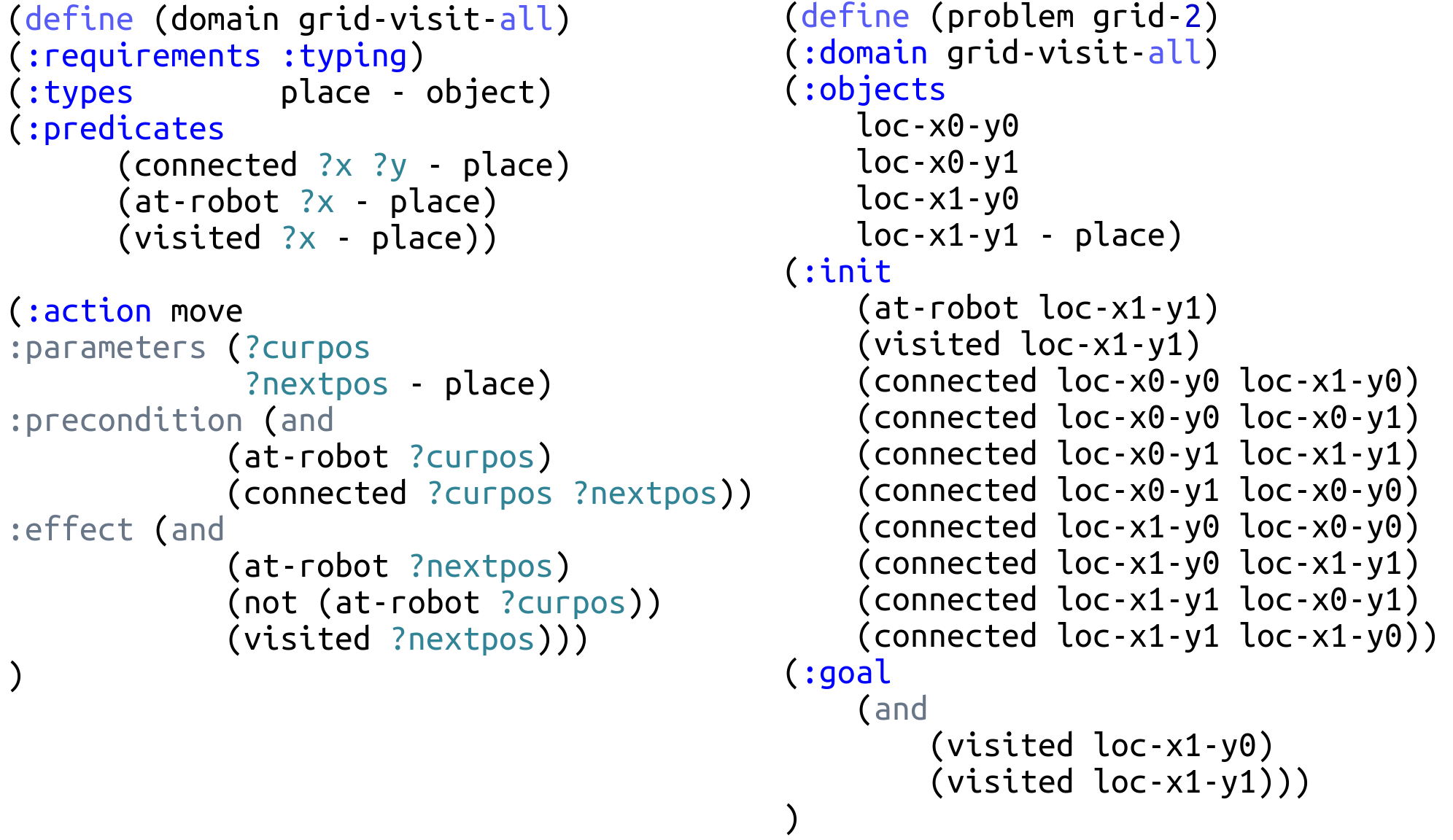}}\hfill
\subfigure[Grounded graph]{\includegraphics[height=1.5in, angle=90]{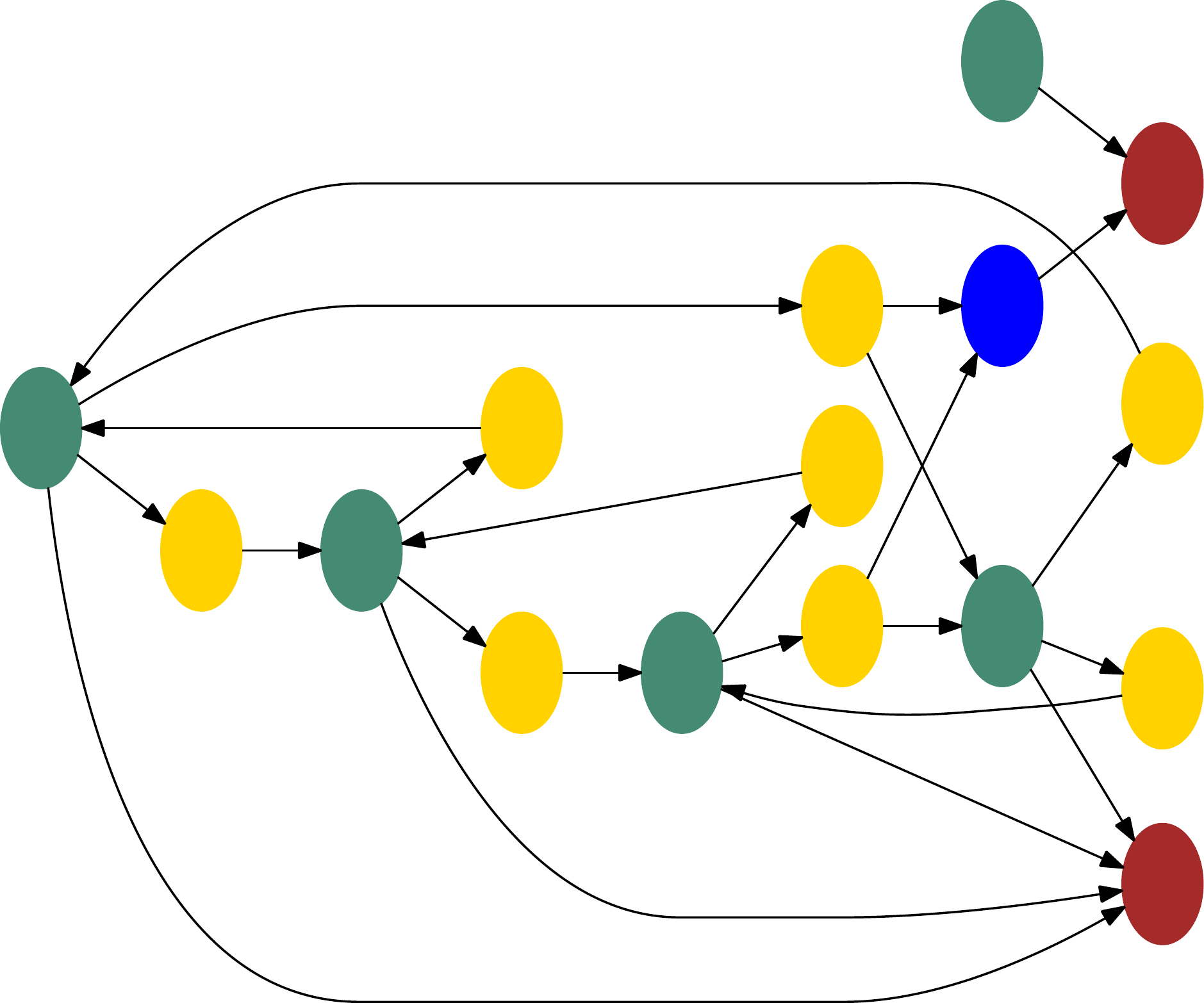}}\hfill
\subfigure[Lifted graph]{\includegraphics[height=1.3in]{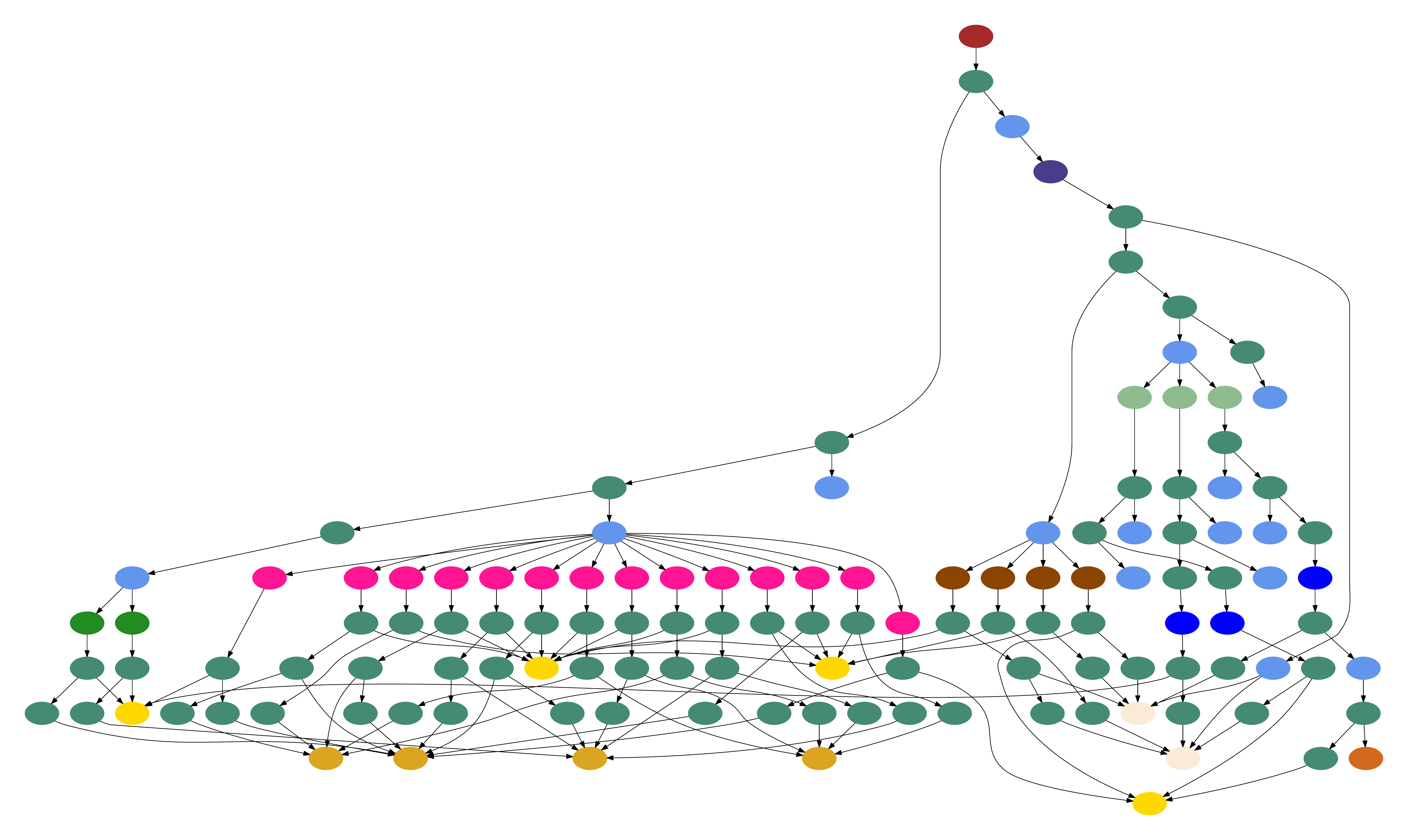}}
\caption{An example planning task (described by using PDDL) and the constructed graphs.}
\label{fig:example}
\end{figure*}

\newcommand{\sasp}{\ensuremath{\textup{SAS}^+{}}}

\newcommand{\lang}{\ensuremath{\mathcal L}}
\newcommand{\ops}{\ensuremath{\mathcal O}}
\newcommand{\axs}{\ensuremath{\mathcal A}}
\newcommand{\cost}{\ensuremath{\textit{cost}}}
\newcommand{\pre}{\ensuremath{\textit{pre}}}
\newcommand{\cond}{\ensuremath{\textit{cond}}}
\newcommand{\eff}{\ensuremath{\textit{eff}}}
\newcommand{\body}{\ensuremath{\textit{body}}}
\newcommand{\head}{\ensuremath{\textit{head}}}
\newcommand{\derived}[1]{\ensuremath{\llbracket#1\rrbracket}}
\newcommand{\init}{\ensuremath{s_0}}
\newcommand{\goal}{\ensuremath{s_\star}}
\newcommand{\variables}{\ensuremath{\mathcal V}}
\newcommand{\dvariables}{\ensuremath{\mathcal V}_{d}}
\newcommand{\dvalues}{\ensuremath{s_d}}
\newcommand{\dom}{\ensuremath{\textit{dom}}}
\newcommand{\vars}{\ensuremath{\textit{vars}}}
\newcommand{\effs}{\ensuremath{\textit{effs}}}
\newcommand{\var}{\ensuremath{\textit{var}}}
\newcommand{\val}{\ensuremath{\textit{val}}}
\newcommand{\inlinecite}[1]{\citeauthor{#1} \citeyear{#1}}
\newcommand{\rawcite}[1]{\citeauthor{#1}, \citeyear{#1}}

\newcommand{\as}{\ensuremath{A}}
\newcommand{\symmetrygraph}{\ensuremath{\textit{ASG}}}

\newtheorem{definition}{Definition}

PDDL tasks are defined over a first-order language \lang\ that consists of
predicates, functions, a set of natural numbers, variables, and constants. Given
$\lang$, a \emph{normalized} \cite{helmert-aij2009} PDDL task is a tuple $\Pi =
\langle \ops, \axs, I, G \rangle$ of \emph{schematic operators}, \emph{schematic
axioms}, \emph{initial state specification}, and \emph{goal specification}, in
so-called \emph{lifted} representation.

Most tools for solving the planning problems (a.k.a. planners) perform grounding as
a first step, followed by translation into a \sasp\ language
\cite{backstrom-nebel-compint1995}. In \sasp, a task $\Pi = \langle \variables,
\ops, \axs, \init, \goal \rangle$ consists of finite-domain \emph{variables},
\emph{ground operators}, \emph{ground axioms}, \emph{initial state}, and
\emph{the goal}.

Our data set consists of two graphical representations per planning task. These
representations losslessly encode the information in the planning task and are
often used for the computation of \emph{structural symmetries}
\cite{shleyfman-et-al-aaai2015}. The graph obtained from the grounded
representation \sasp\ is called the \emph{problem description graph (PDG)}
\cite{pochter-et-al-aaai2011}. We present here the definition extended to
support conditional effects and axioms and refer the reader to
\citet{sievers-et-al-aaai2019} for further details. 

\begin{definition}
    Let $\Pi = \langle \variables, \dvariables, \ops, \axs, \dvalues, \init, \goal
    \rangle$ be a \sasp\ task.
    The \emph{problem description graph} of $\Pi$ is the digraph
    $\langle N, E\rangle$ with nodes
    \[
      N = \{n_0, n_\star\}
        \cup \{n_v \mid v \in \variables\} \cup N_f \cup N_\ops \cup \{n_a \mid a \in \axs \},
    \]
    where $N_f = \{n_v^d \mid v \in \variables, d \in \dom(v)\}$ and $N_\ops =
    \{n_o \mid o \in \ops\} \cup  \{  n_o^e \mid o \in \ops, e \in \effs(o)\}$,
    and edges
    \[
      E = E_0 \cup E_\star \cup E_v \cup E_a \cup E_o, \text{ where}
    \]
    \[
    \begin{array}{r@{~}c@{~}l}
      E_0 &=& \{\langle n_0, n_v^d \rangle \mid \init[v] = d\} \\
      E_\star &=& \{\langle n_g, n_v^d \rangle \mid v \in \vars(\goal), \goal[v] = d\}\\
      E_v &=& \{\langle n_v, n_v^d \rangle \mid d \in \dom(v)\}\\
      E_a &=& \{\langle n_a, n_v^d \rangle \mid a \in \axs, v \in \vars(\pre(a)), \pre(a)[v] = d\}\\
      & \cup & \{\langle n_a, n_v^d \rangle \mid a \in \axs, \var(a) = v, val(a) = d\}\\
      E_o &=& \{\langle n_o, n_v^d \rangle \mid o \in \ops, v \in \vars(\pre(o)), \pre(o)[v] = d\}\\
      & \cup & \{\langle n_o, n_o^e \rangle \mid e \in \effs(o)\}\\
      & \cup & \{\langle n_v^d, n_o^e \rangle \mid \langle c,\cdot,\cdot\rangle
      \in \effs(o), v \in \vars(c), c[v] = d\}\\
      & \cup & \{\langle n_o^e, n_v^d \rangle \mid \langle \cdot,v,d\rangle \in
      \effs(o)\}.\\
    \end{array}
    \]
  \end{definition}
  
The graph obtained from the lifted PDDL representations is called the
\emph{abstract structure graph (ASG)} \cite{sievers-et-al-icaps2019}. Planning
tasks in PDDL can be naturally modeled as abstract structures, which, in turn,
can be represented as graphs. In what follows we present the definitions of
\emph{abstract structures} and \emph{abstract structure graphs}, referring the
reader to \citet{sievers-et-al-icaps2019} for further details.

\begin{definition}[\rawcite{sievers-et-al-icaps2019}]
  Let $S$ be a set of symbols, where each $s\in S$ is associated with a
  \emph{type} $t(s)$. The set of \emph{abstract structures} over $S$ is
  inductively defined as follows:
  \begin{itemize}
    \item each symbol $s \in S$ is an abstract structure, and

    \item for abstract structures $\as_1, \dots, \as_n$, the set $\{
    \as_1, \dots, \as_n \}$ and the tuple $\langle \as_1, \dots, \as_n \rangle$
    are abstract structures.
  \end{itemize}
\end{definition}

Using the language \lang\ of a PDDL task $\Pi$, each part of $\Pi$ can
inductively be defined as an abstract structure, with the symbols of \lang\
forming the basic abstract structures. Finally, abstract structures can be
naturally turned into a graph.

\begin{definition}[\rawcite{sievers-et-al-icaps2019}]
  Let \as\ be an abstract structure over $S$. The \emph{abstract structure
  graph} $\symmetrygraph_\as$ is a digraph $\langle N, E \rangle$,
  defined as follows.
  \begin{itemize}
    \item $N$ contains a node ${\as}$ for the abstract structure $\as$. If
    $N$ contains a node for $\as'= \{ \as_1, \dots, \as_n \}$ or $\as'= \langle
    \as_1, \dots, \as_n \rangle$, it also contains the nodes for ${\as_1},\dots,
    {\as_n}$.

    \item For every set (sub-)structure $A' = \{\as_1,\dots,\as_n\}$ there are edges
    $A'\rightarrow \as_i$ for $i\in\{1,\dots,n\}$.

    \item For every tuple (sub-)structure $A' = \langle \as_1, \dots, \as_n\rangle$,
    the graph contains auxiliary nodes $n^{A'}_1,\dots, n^{A'}_n$, an edge $A'
    \rightarrow n^{A'}_1$, and edges $n^{A'}_{i-1}\rightarrow
    n^{A'}_i$ for $1 < i \leq n$. For each component $\as_i$, there is an edge $n^{A'}_i\rightarrow
    \as_i$.
  \end{itemize}
\end{definition}

Note that the edges in ASGs are from the abstract structures to their sub-structures, which results in acyclic graphs.

In both PDG and ASG, the node features are one-hot according to the self-explanatory node type indicated in the above definitions.

The aim in classical planning is to find a sequence of ground operators that, if
applied to the initial state, will necessarily transform it into a goal state.
Such a sequence is called a \emph{plan}. Assigning a quantitative cost to each
ground operator, the cost of a plan is defined as the sum over the costs of its
operators. The goal of cost-optimal classical planning is to find a provably
cheapest plan.
There exist dozens if not hundreds of highly parameterized methods
for heuristic guidance computation, giving rise to an enormous possible number
of planners. As even the classical planning is PSPACE-hard, there cannot be one
planner that will work well on all possible planning problems. Thus, finding a
planner that works well on a given planning problem is a challenging task.

While planners are often domain-independent, in the sense that they depend only
on the information encoded in PDDL, the planning tasks encode computational
problems from various domains. These domains range from puzzles or one-person
games (e.g., towers of Hanoi, 15-puzzle, freecell, and sokoban), to real-life
domains (e.g., task planning and automated control of autonomous systems:
greenhouse logistics, rovers, elevators, satellites), as well as emerging domains (e.g.,
genome editing distance computation). 

Many of the existing domains were introduced through International Planning
Competitions, which were held  regularly since 1998. Each such
competition, intended for comparing the performance of domain-independent
planners, introduced new, previously unseen domains on which submitted planners were
tested. In many cases, the authors of the domains supplied not only the
planning tasks, but also the generator that allowed for creating additional
tasks. Some of these generators can be found at, e.g.,
\url{https://bitbucket.org/planning-researchers/pddl-generators}.
Hence, these generators may be used to extend the current data set with little effort, although for benchmarking purpose we did not include any such task in the data set.



\section{Statistics}\label{sec:stat}
A number of graph statistics, compared with those of commonly used datasets~\cite{Kersting2016} for benchmarking graph kernels and graph neural networks, are reported in Table~\ref{tab:stat} and Figures~\ref{fig:stat} and~\ref{fig:stat2} in the supplementary material. Observations follow.

\begin{enumerate}[leftmargin=*]
\item \textbf{The IPC graphs are significantly larger.} The graphs in other data sets under comparison generally have tens to hundreds of nodes, but 39\% of the graphs in IPC-grounded and 63\% in IPC-lifted have over 1,000 nodes. The largest graph in IPC-grounded has 87,140 nodes, and the number for IPC-lifted is 238,909.

\item \textbf{Note that the size of the largest graph is often the memory bottleneck indicator for graph neural networks, because the batch size is at least this number in stochastic training.} Hence, our data set poses substantial challenges for the computation of many neural graph models.

\item \textbf{The sizes of the IPC graphs are highly skewed, compared to those of other data sets.} For many machine learning tasks, especially in the unsupervised setting, the notion of similarity is key to clustering and categorization. When the sizes of two graphs significantly differ, the intuition of similarity is challenged. After all, what does it mean by saying ``a graph with 10 nodes is similar to another graph with 100,000 nodes?''

\item \textbf{The lifted graphs are the most sparse, compared to the grounded ones and graphs in other data sets.}

\item \textbf{Similar to many other data sets, the IPC graphs are not necessarily connected.} However, the main connected component generally dominates. Hence, graph neural networks still suffer the memory bottleneck caused by the exceedingly large graphs.

\item \textbf{Despite the difference in size and density, the IPC graphs have a moderate diameter, similar to other data sets.} The number of layers in a graph neural network of neighborhood-aggregation style is often questioned beyond hyperparameter tuning; and speculation attributes to the diameter of the graphs. Meanwhile, it has been widely acknowledged that neighborhood aggregation is a type of Laplacian smoothing and too many layers lead to oversmoothing~\citep{Li2018,Xu2018,Klicpera2019}. The diameter statistics may be useful for the analysis of the role of small-world structures handled by graph neural networks.
\end{enumerate}

\begin{table*}[ht]
\centering
\begin{threeparttable}
\caption{Statistics of IPC, compared with that of the other commonly used benchmark data sets.}
\label{tab:stat}
\centering
\begin{tabular}{ccccc}
\hline
& IPC-grounded & IPC-lifted & REDDIT-MULTI-12k & REDDIT-BINARY \\
\hline
Type          & directed & DAG & undirected & undirected \\
\#Graphs      & 2,439 & 2,439 & 11,929 & 2,000 \\
Total \#Nodes & 6,233,856 & 9,816,948 & 4,669,116 & 859,254 \\
Max \#Nodes   & 87,140 & 238,909 & 3,782 & 3,782 \\
Mean (Std) \#Nodes
              & 2,555.9 (6,099.0) & 4,025.0 (14,507.6) & 391.4 (428.7) & 429.6 (554.1) \\
Mean (Std) Ave Degree\tnote{1}
              & 12.3 (131.0) & 2.9 (35.1) & 4.7 (27.6) & 4.6 (41.3) \\
Mean (Std) \#CC\tnote{2}
              & 1.09 (0.61) & 1.14 (0.49) & 2.81 (2.65) & 2.48 (2.47) \\
Mean (Std) Diam\tnote{3,4}
              & 8.2 (2.3) & 17.1 (1.5) & 10.9 (3.1) & 9.7 (3.1) \\
\hline
\end{tabular}

\begin{tablenotes}
\item [1] ``Ave Degree'' is the average node degree (of the undirected version of the graph).
\item [2] ``CC'' means connected components (of the undirected version of the graph).
\item [3] ``Diam'' means diameter. Because a graph may consist of multiple connected components, we define the diameter as the maximum of the diameters of each connected component.
\item [4] For large graphs, the diameter is too costly to compute. Hence, for IPC, only the diameters of 94.3\% of the graphs are computed. For other data sets, diameters of all graphs are computed.
\end{tablenotes}

\vskip 10pt
\begin{tabular}{ccccccc}
\hline
 & COLLAB & NCI1 & DD & PROTEINS & ENZYMES & MUTAG \\
\hline
Type          & undirected & undirected & undirected & undirected & undirected & undirected\\
\#Graphs      & 5,000 & 4,110 & 1,178 & 1,113 & 600 & 188\\
Total \#Nodes & 372,474 & 122,747 & 334,925 & 43,471 & 19,580 & 3,371\\
Max \#Nodes   & 492 & 111 & 5,748 & 620 & 126 & 28\\
Mean (Std) \#Nodes
              & 74.5 (62.3) & 29.9 (13.6) & 106.5 (284.3) & 39.1 (45.8) & 32.6 (15.3) & 18.0 (4.6)\\
Mean (Std) Ave Degree\tnote{1}
              & 132.0 (158.5) & 4.3 (1.6) & 10.1 (3.4) & 7.5 (2.3) & 7.6 (2.3) & 4.4 (1.5)\\
Mean (Std) \#CC\tnote{2}
              & 1 (0) & 1.19 (0.57) & 1.02 (0.18) & 1.08 (0.52) & 1.24 (3.61) & 1 (0)\\
Mean (Std) Diam\tnote{3,4}
              & 1.9 (0.3) & 13.3 (5.1) & 19.9 (7.7) & 11.6 (7.9) & 10.9 (4.8) & 8.2 (1.8)\\
\hline
\end{tabular}

\end{threeparttable}
\end{table*}

\section{Example Use}\label{sec:use}
For an illustration of the use of the data set, we focus on the problem of cost-optimal planning, whose goal is to solve as many tasks by using cost-optimal planners as possible, each given a time limit $T$. Hence, for each of the 17 target values, we convert it to 0 if the value $\le T$ and 1 otherwise. For each target, the problem becomes a binary classification and thus a probability value between 0 and 1 is output. We select the planner corresponding to the smallest probability and confirm success if its actual planning time is smaller than the timeout limit $T$. Test accuracy (percentage of successfully solved tasks) is reported.

Three methods for comparison are (a) an image-based CNN whereby the gray-scale image is converted from the adjacency matrix of the graph; (b) a graph convolutional network (GCN) \citep{Kipf2017} with attention readout; and (c) a gated graph neural network (GG-NN) \citep{Li2016}. For details of the CNN architecture, see~\citet{katz-et-al-ipc2018}.


The data set has been pre-split for training, validation, and testing. Table~\ref{tab:planning.results} reports the test accuracy. Additionally, we re-split the training/validation combination as a form of cross validation, whereby we fix the test set because it comes from the most recent International Planning Competition. Two forms of random re-splits are possible. One is to preserve the domains of the planning tasks (i.e., tasks from the same domain cannot appear in both training and validation), and the other is free from this restriction. We call the former \emph{domain split} and the latter \emph{random split}. For each type of re-split, we perform ten randomizations. Table~\ref{tab:planning.results2} reports the test accuracy together with standard deviation. From both tables, one sees that the lifted graphs yield much higher accuracy and GCN outperforms the other two methods.

\begin{table}[ht]
\centering
\caption{Percentage of solved tasks in the test set.}
\label{tab:planning.results}
\begin{tabular}{ccc}
\hline
Method & Grounded & Lifted \\
\hline
CNN   & 73.1\% & 86.9\% \\
GCN   & 80.7\% & 87.6\% \\
GG-NN & 77.9\% & 81.4\% \\
\hline
\end{tabular}
\end{table}

\begin{table}[ht]
\centering
\caption{Percentage of solved tasks in the test set (lifted graphs). Multiple training/validation splits.}
\label{tab:planning.results2}
\begin{tabular}{ccc}
\hline
Method & Domain Splits & Random Splits\\
\hline
CNN   & 82.1\% (6.6\%) & 86.1\% (5.5\%)\\
GCN   & 85.6\% (5.5\%) & 87.2\% (3.5\%)\\
GG-NN & 76.6\% (5.8\%) & 74.4\% (2.7\%)\\
\hline
\end{tabular}
\end{table}

\section{Conclusions}\label{sec:conclude}
We have described a new data set, IPC, for benchmarking graph-based learning models (e.g., graph kernels and graph neural networks) in classification, regression, and related uses. The graphs are constructed from AI planning tasks appearing in International Planning Competitions, without requiring human efforts for labeling, and may be extended with random instances of planning problems. The data set has distinctively different statistics from other popularly used benchmarks: the graphs are much larger and their sizes vary substantially. Moreover, the lifted version of the data set is comprised of directed acyclic graphs, enabling the development of specialized graph models. We anticipate that the data set is a valuable inclusion to the current collection of commonly used benchmarks for validating the effectiveness of existing and forthcoming graph methods.


\bibliography{abbrv-short,reference2,crossref-short2}
\bibliographystyle{icml2019}

\appendix
\section{Additional Information of Graph Statistics}
\begin{figure*}[ht]
\centering
\subfigure[IPC-grounded]{\includegraphics[width=.19\linewidth]{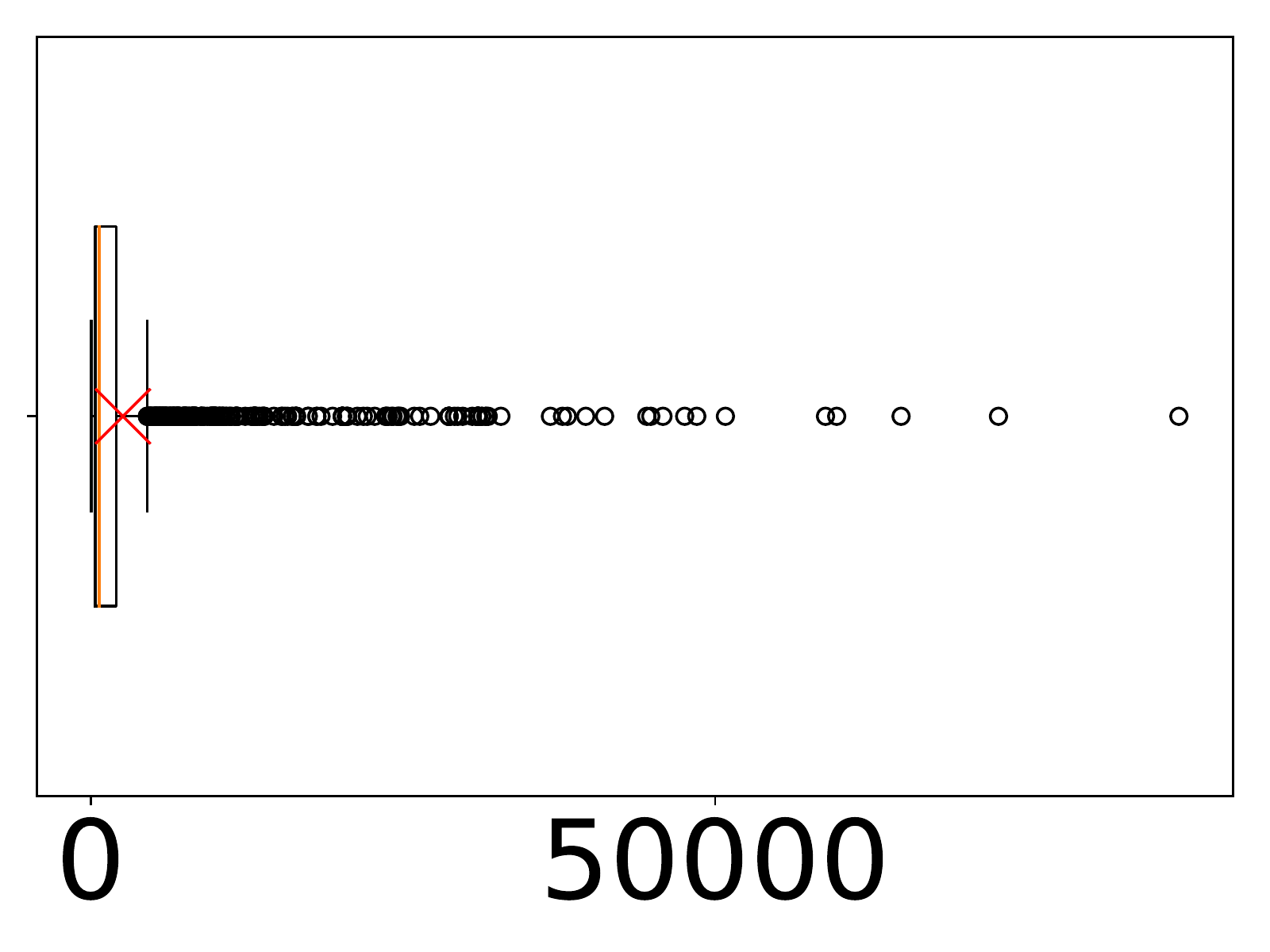}}
\subfigure[IPC-lifted]{\includegraphics[width=.19\linewidth]{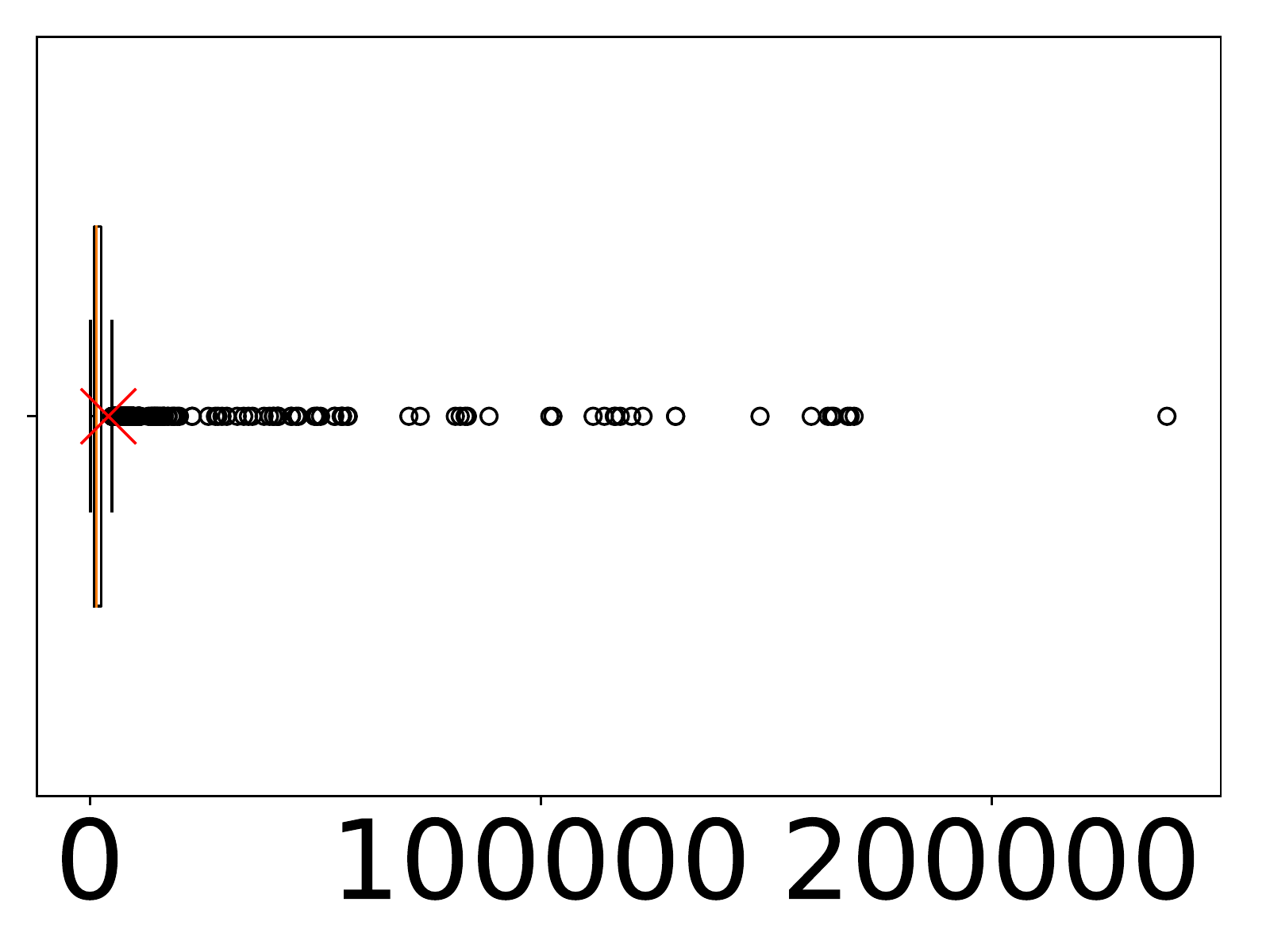}}
\subfigure[REDDIT-MULTI-12k]{\includegraphics[width=.19\linewidth]{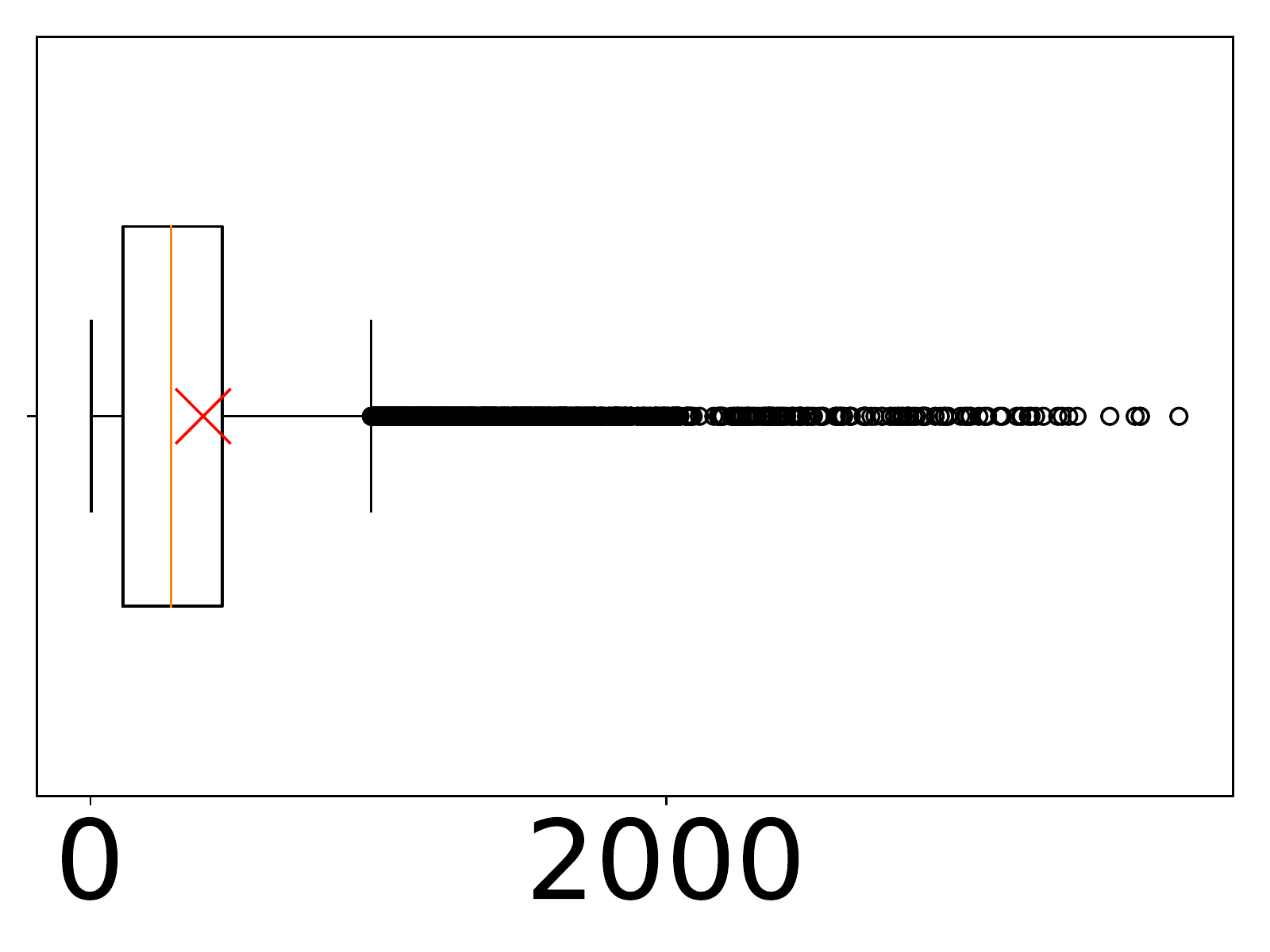}}
\subfigure[REDDIT-BINARY]{\includegraphics[width=.19\linewidth]{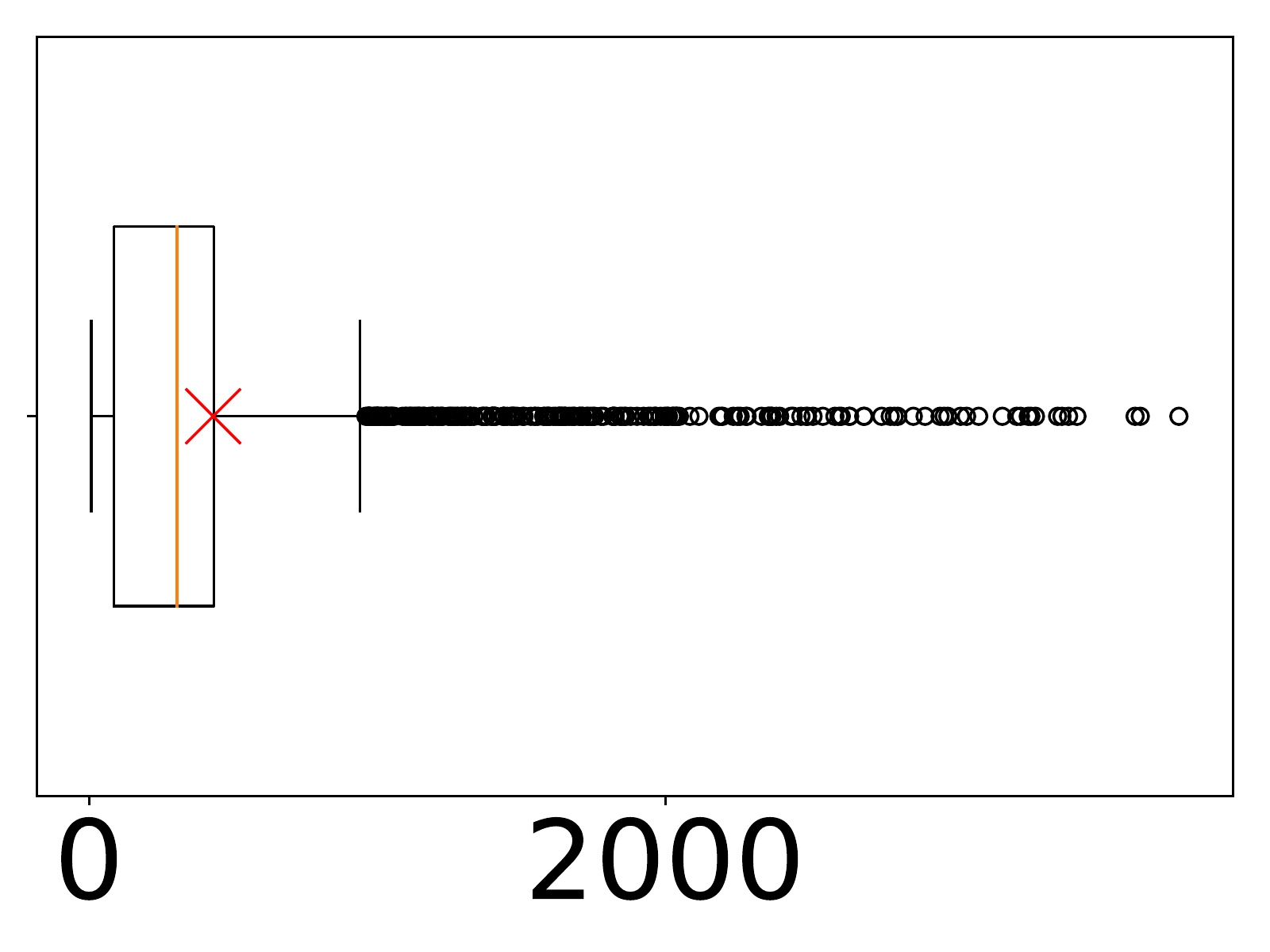}}
\subfigure[COLLAB]{\includegraphics[width=.19\linewidth]{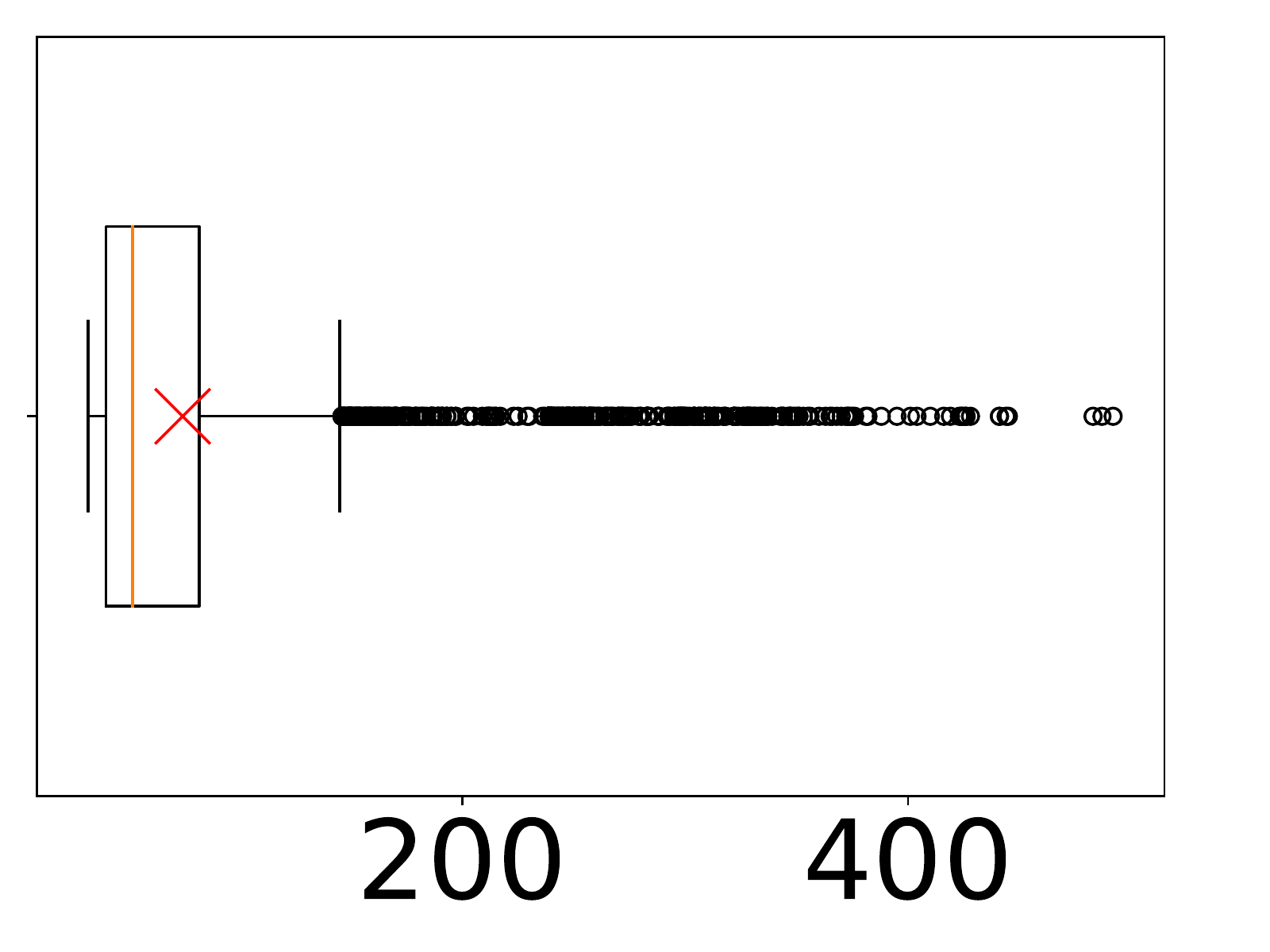}}\\
\subfigure[NCI1]{\includegraphics[width=.19\linewidth]{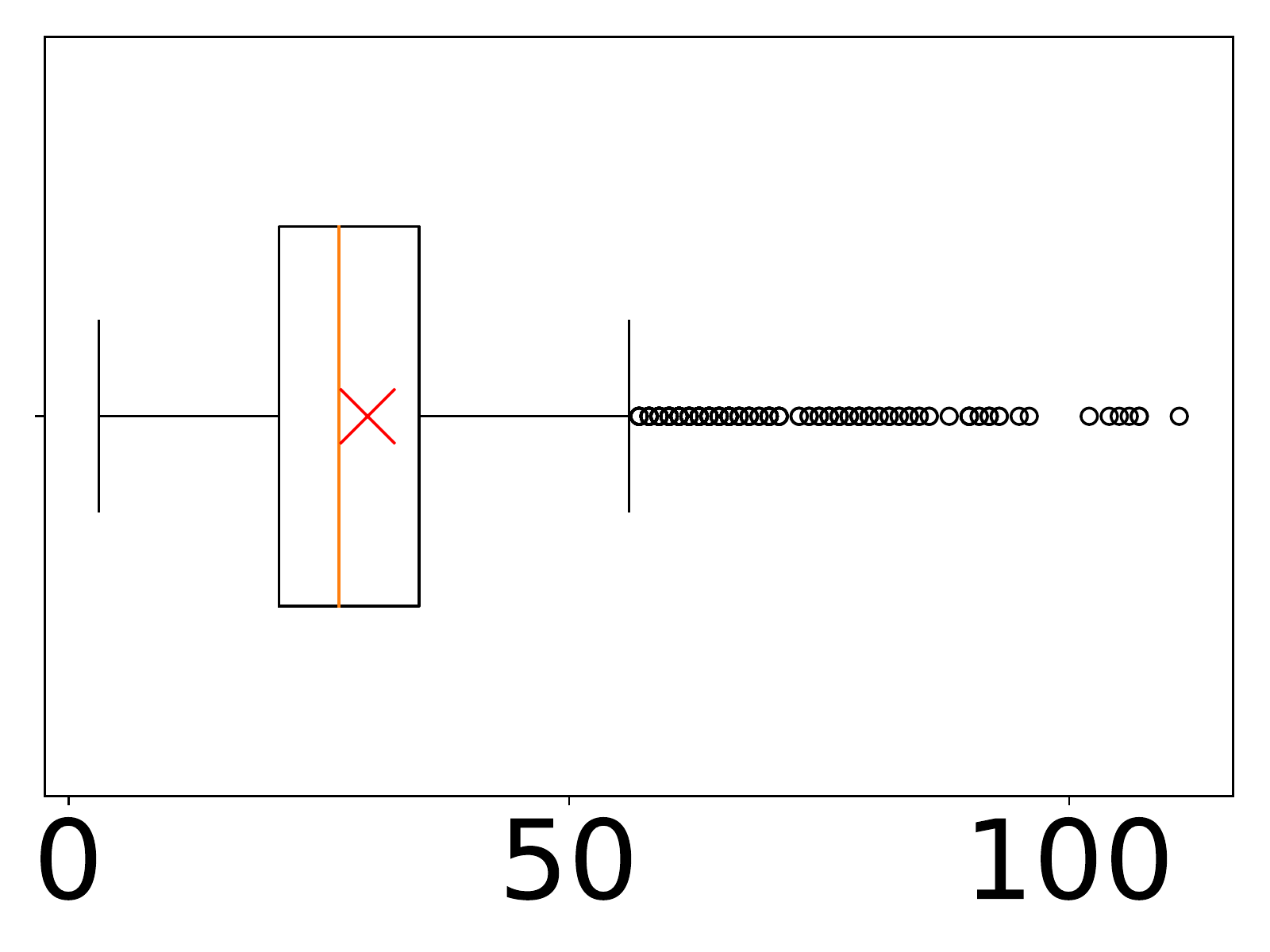}}
\subfigure[DD]{\includegraphics[width=.19\linewidth]{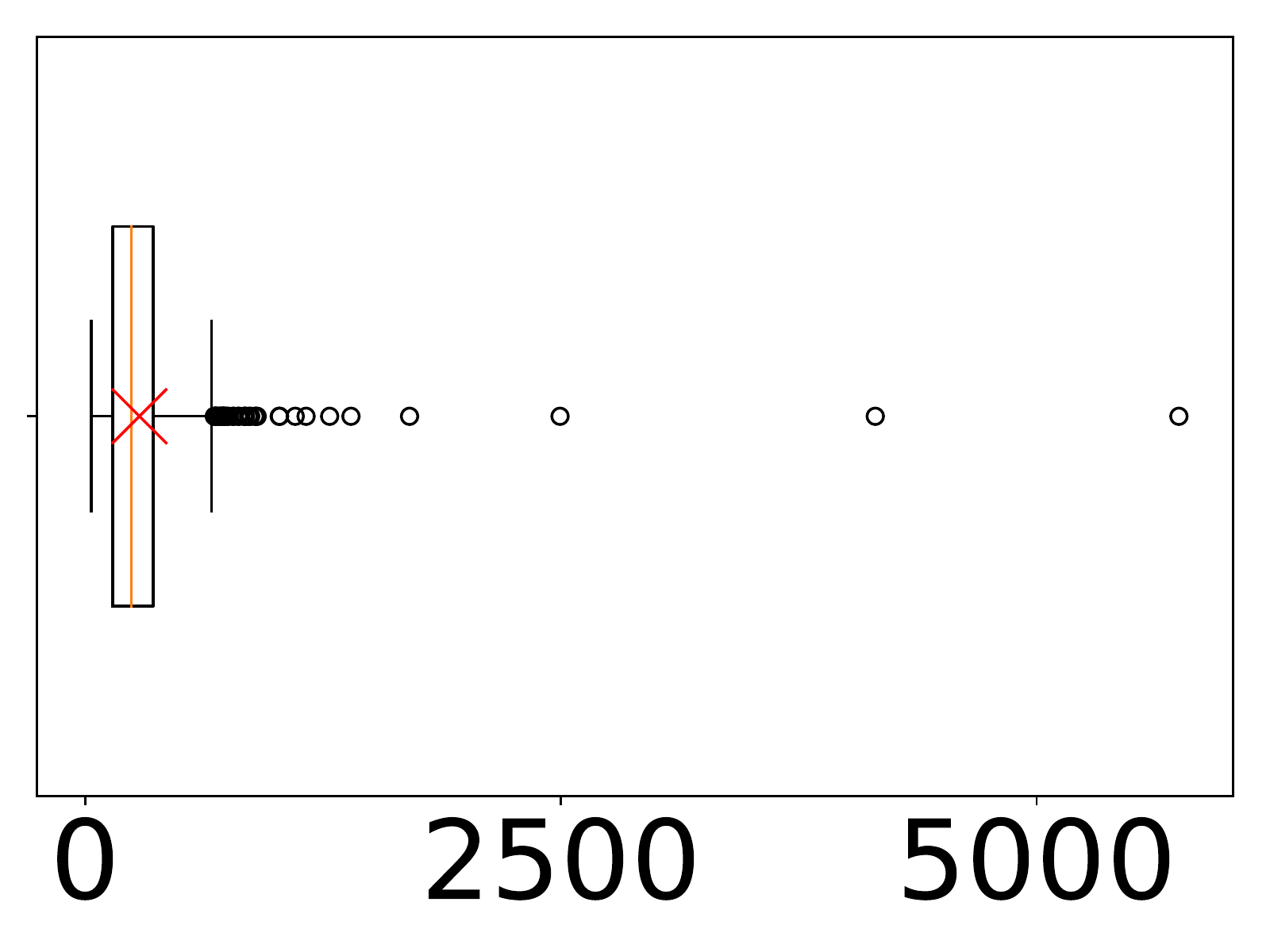}}
\subfigure[PROTEINS]{\includegraphics[width=.19\linewidth]{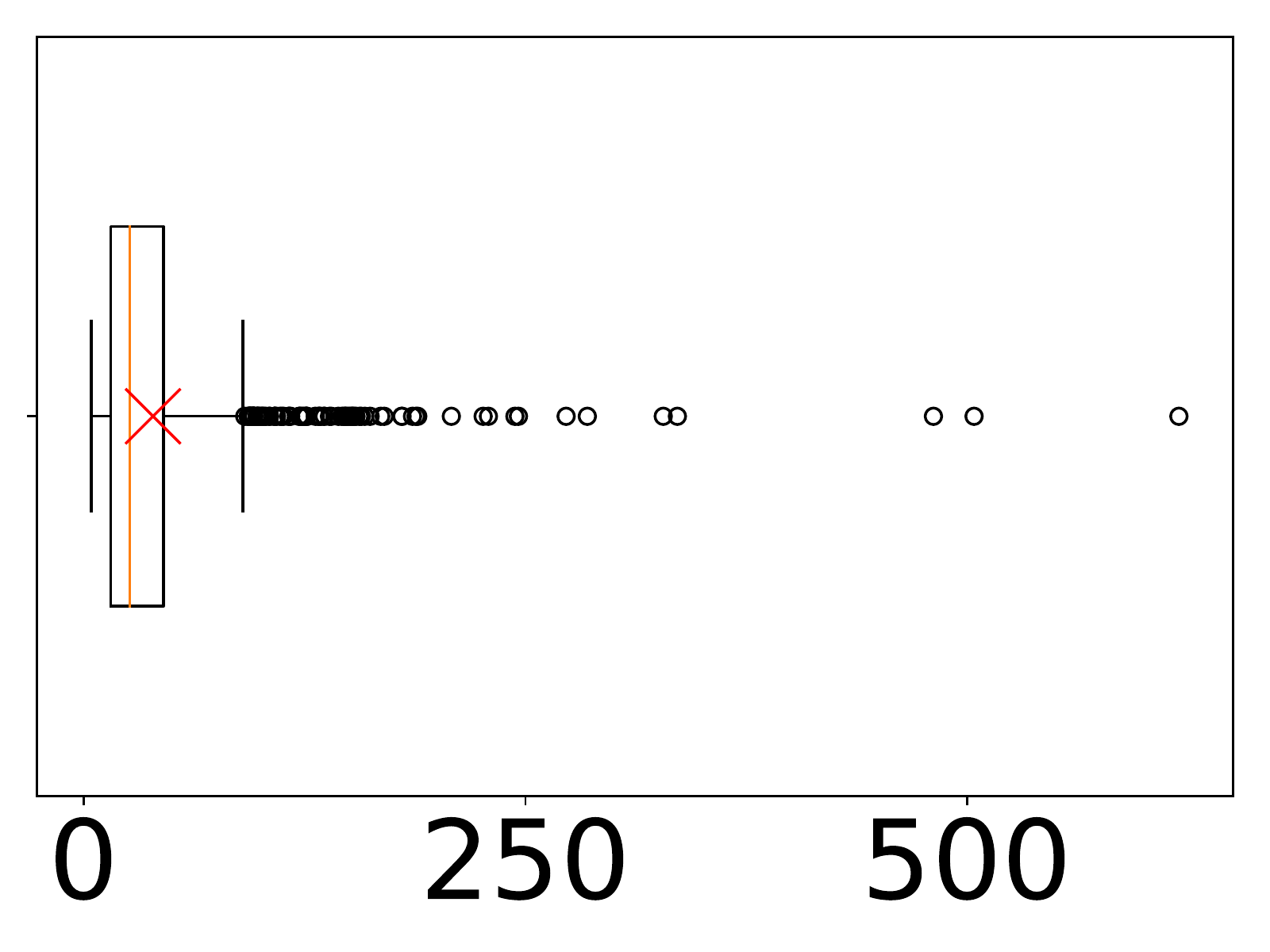}}
\subfigure[ENZYMES]{\includegraphics[width=.19\linewidth]{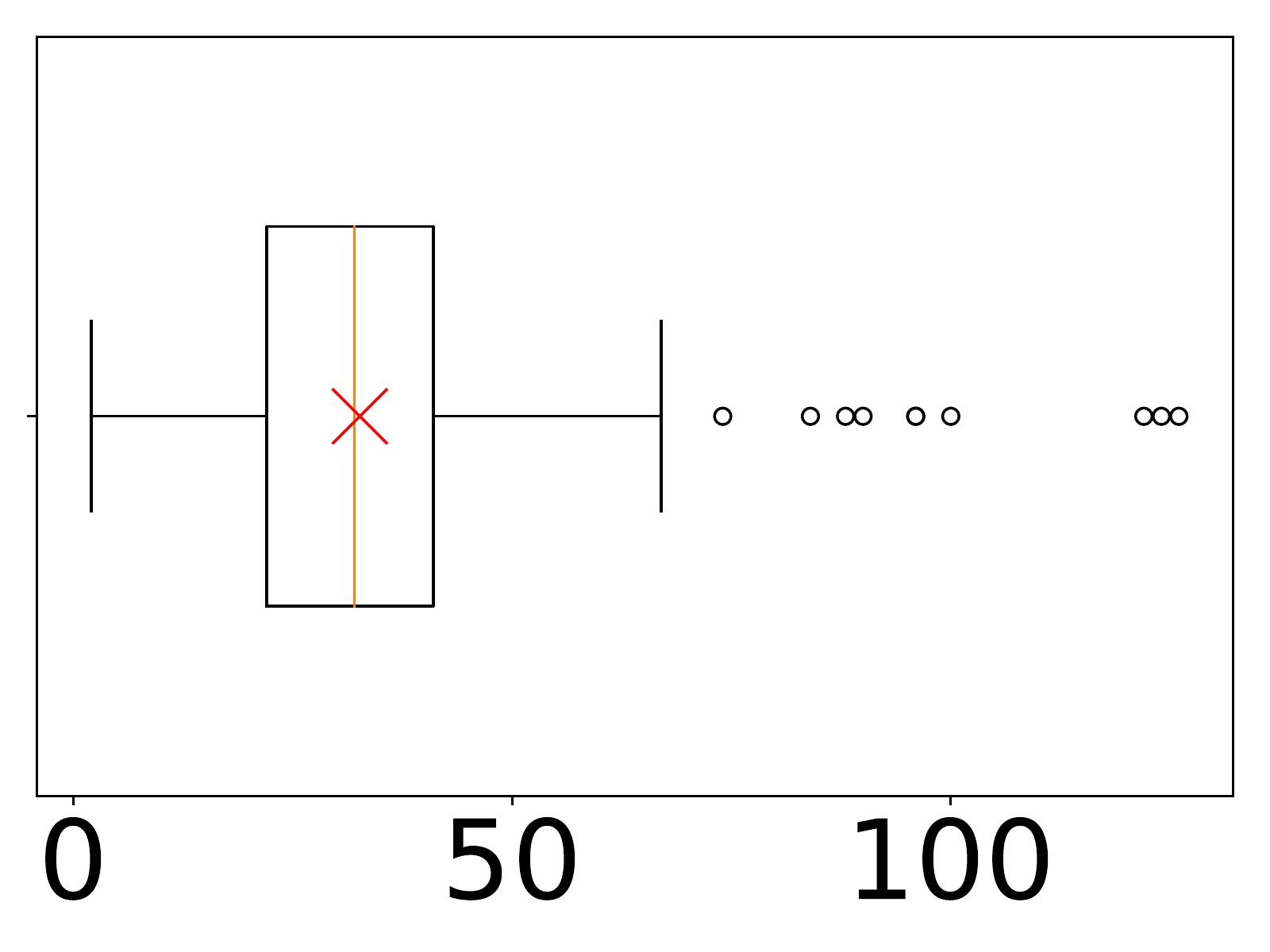}}
\subfigure[MUTAG]{\includegraphics[width=.19\linewidth]{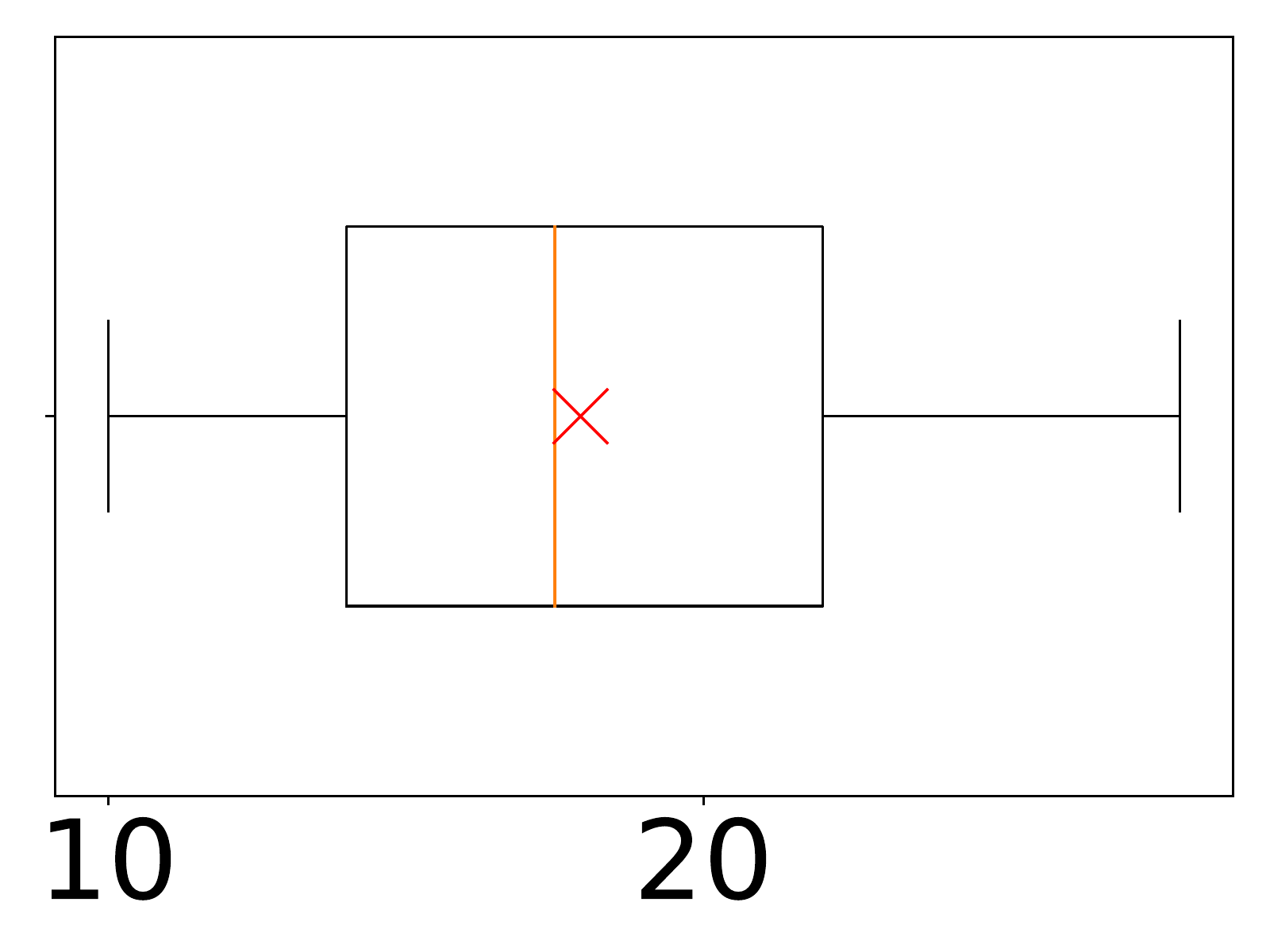}}
\caption{Box plot of the graph size distribution. The orange bar marks the median; the red cross the mean.}
\label{fig:stat}
\end{figure*}

\begin{figure*}[ht]
\centering
\subfigure[IPC-grounded]{\includegraphics[width=.19\linewidth]{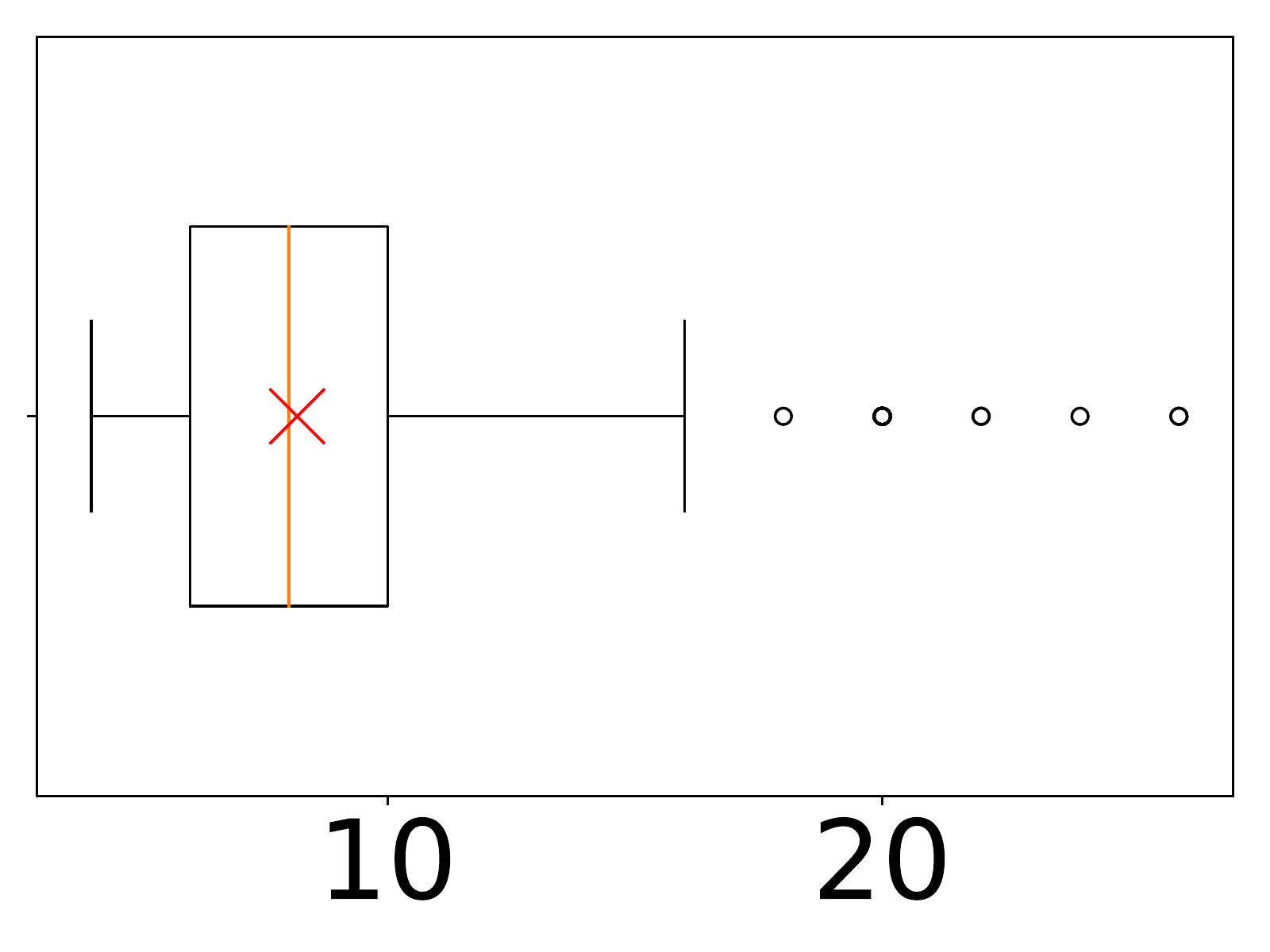}}
\subfigure[IPC-lifted]{\includegraphics[width=.19\linewidth]{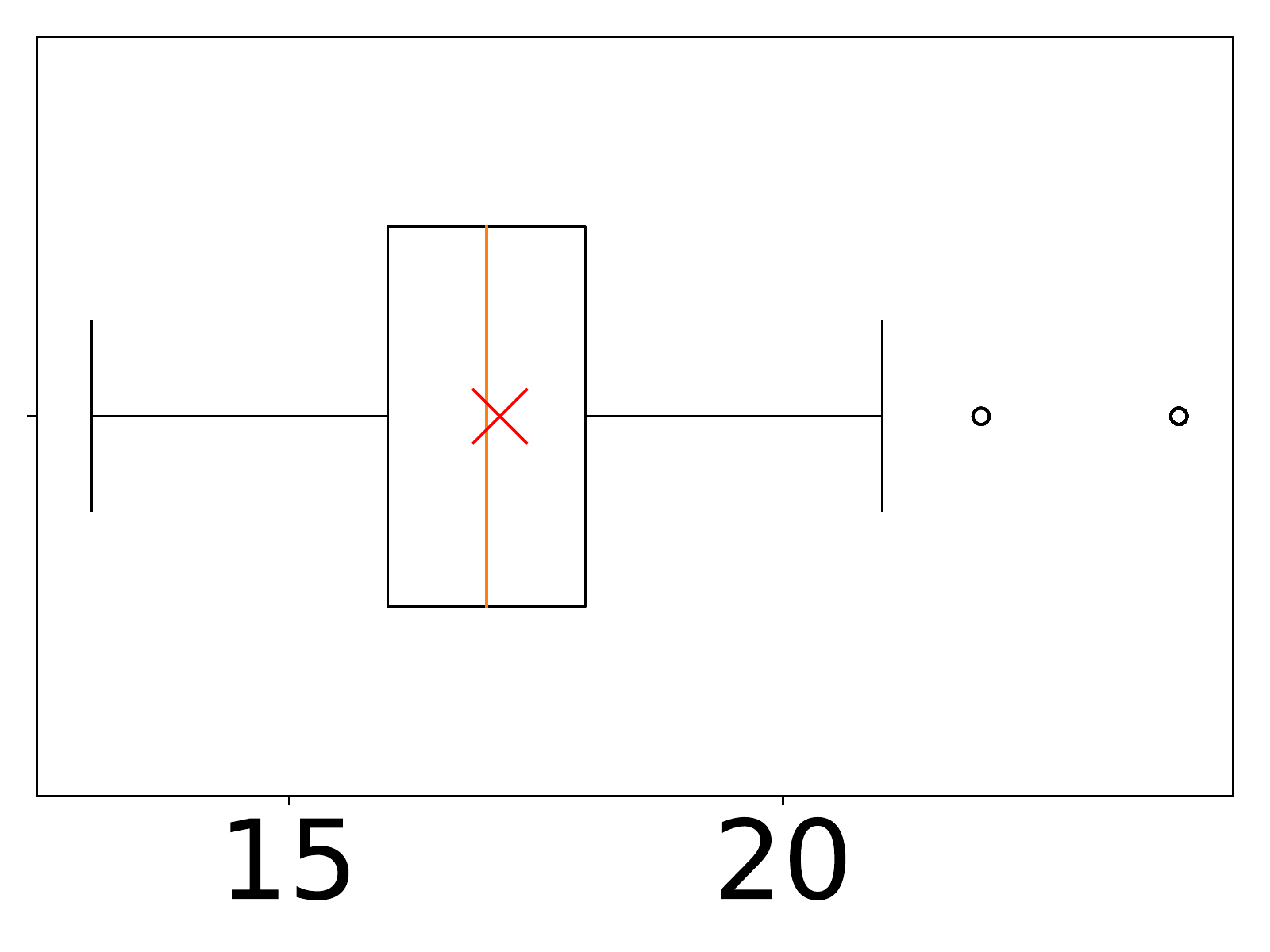}}
\subfigure[REDDIT-MULTI-12k]{\includegraphics[width=.19\linewidth]{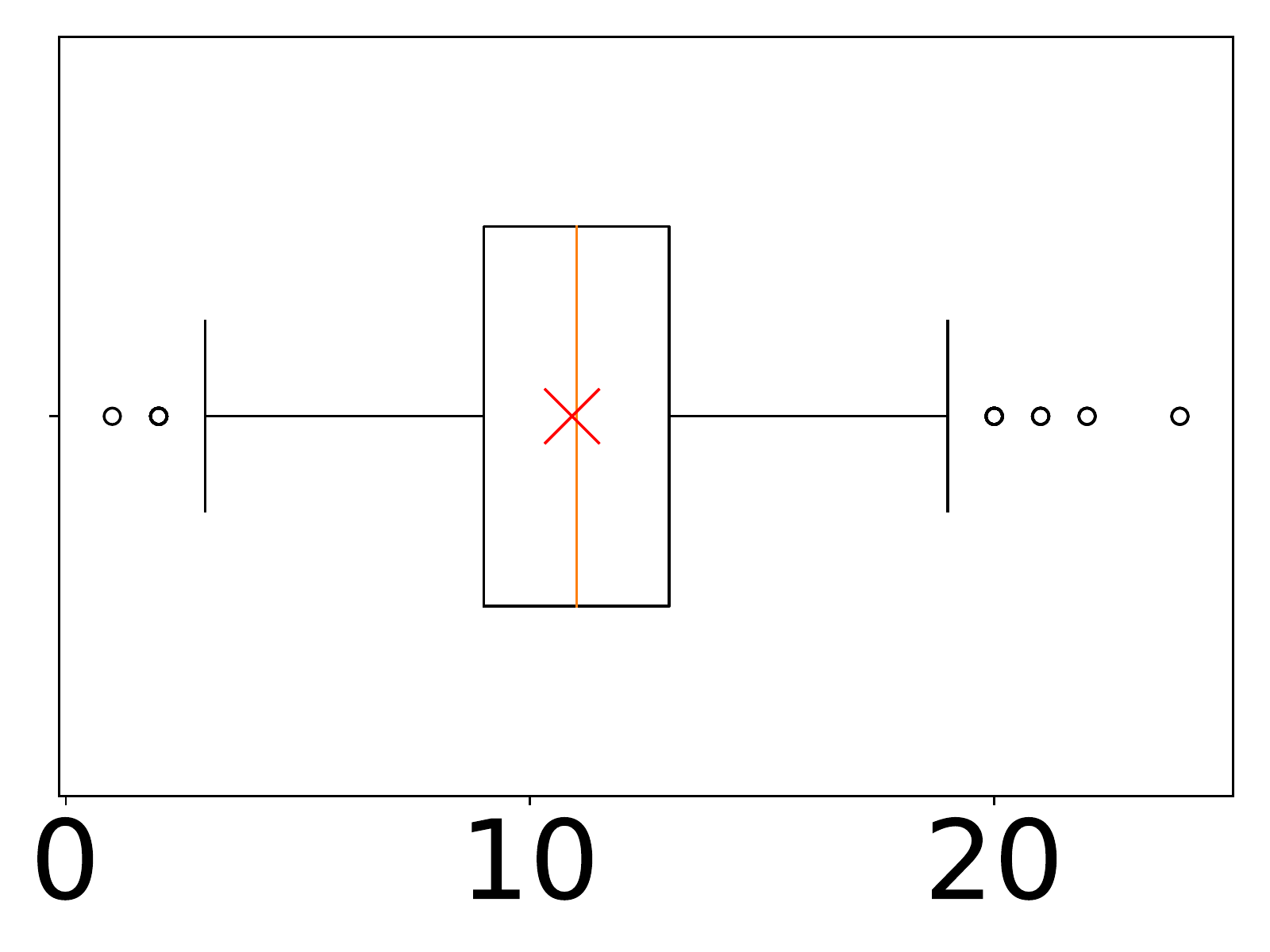}}
\subfigure[REDDIT-BINARY]{\includegraphics[width=.19\linewidth]{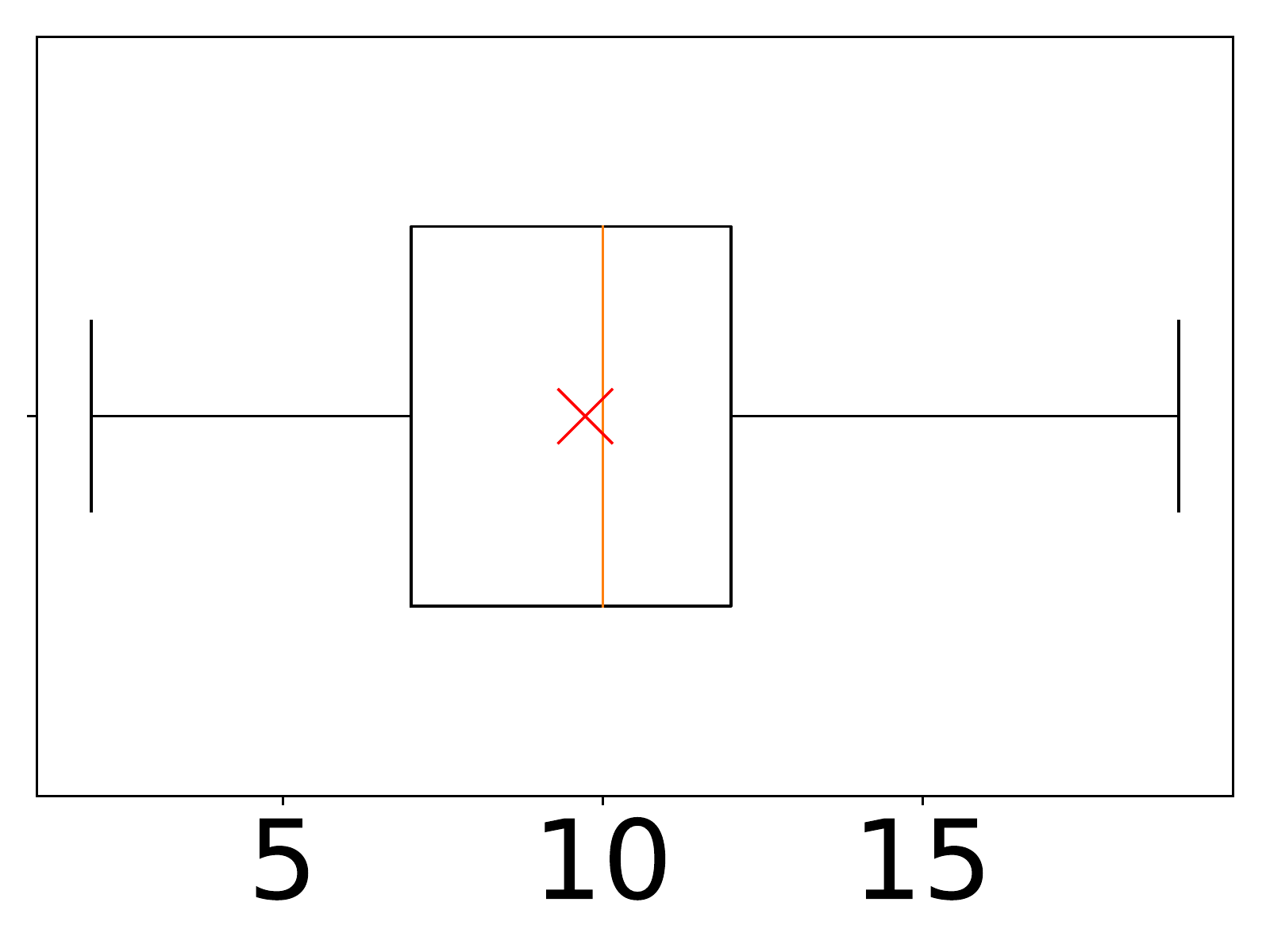}}
\subfigure[COLLAB]{\includegraphics[width=.19\linewidth]{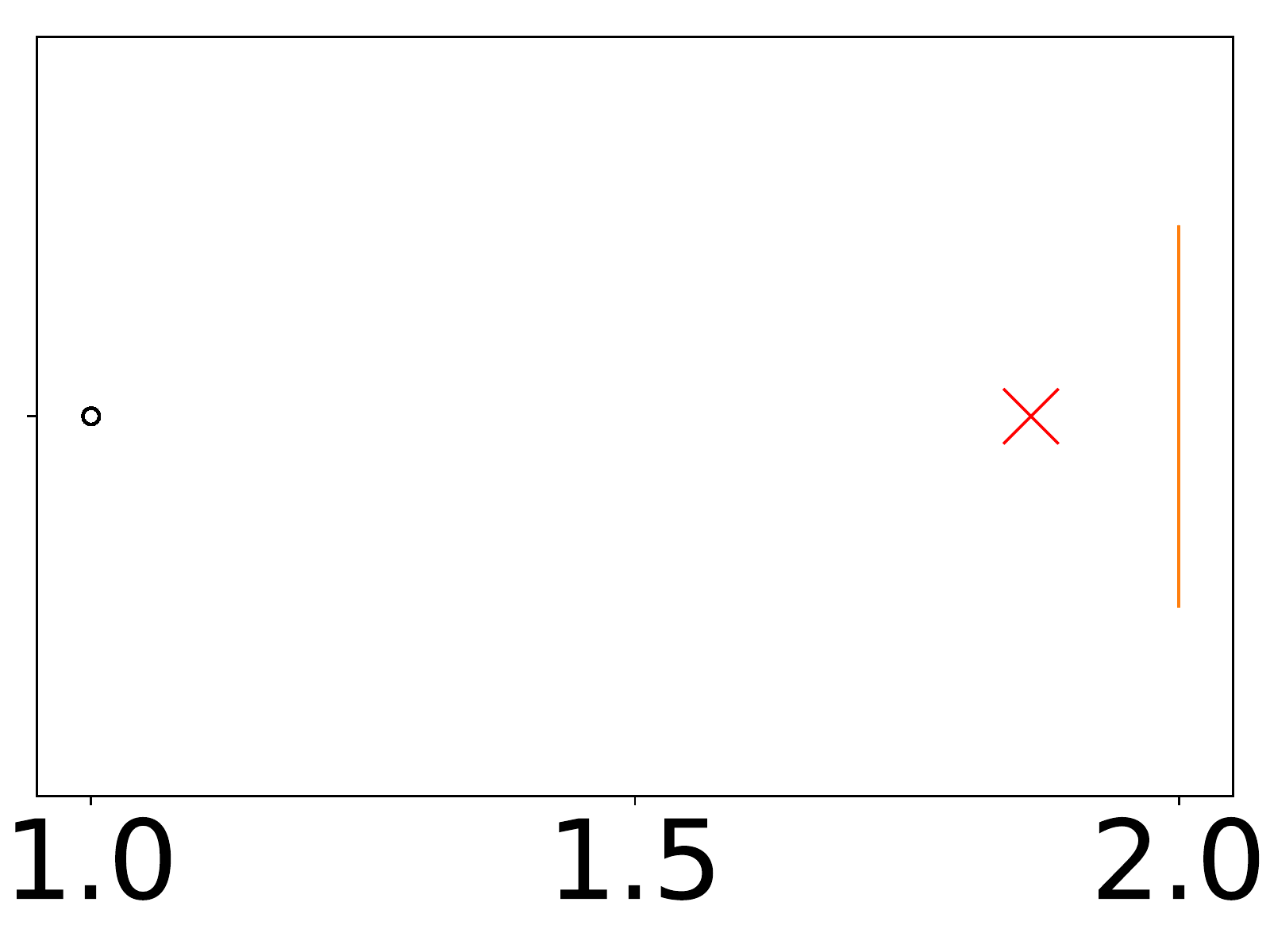}}\\
\subfigure[NCI1]{\includegraphics[width=.19\linewidth]{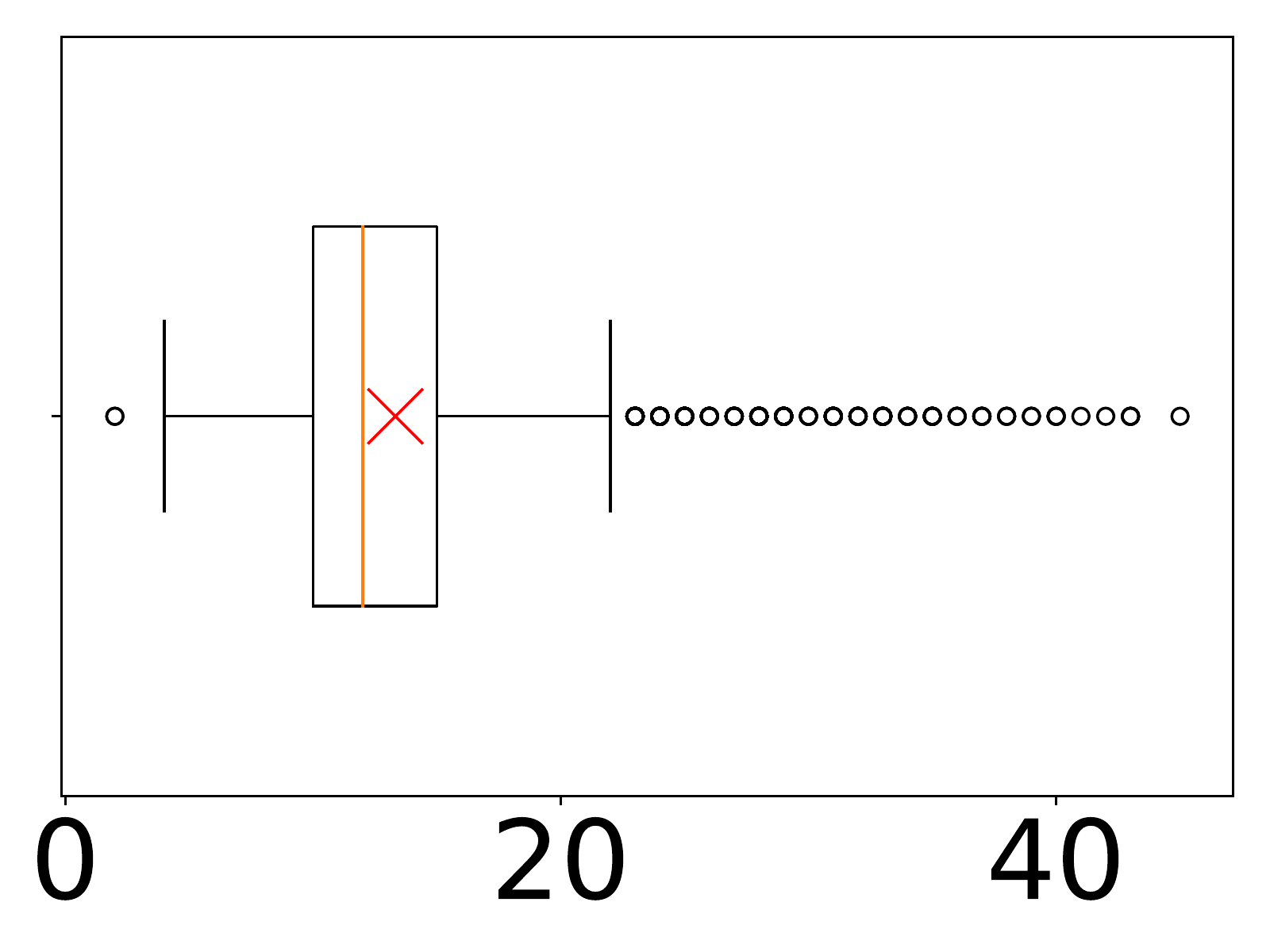}}
\subfigure[DD]{\includegraphics[width=.19\linewidth]{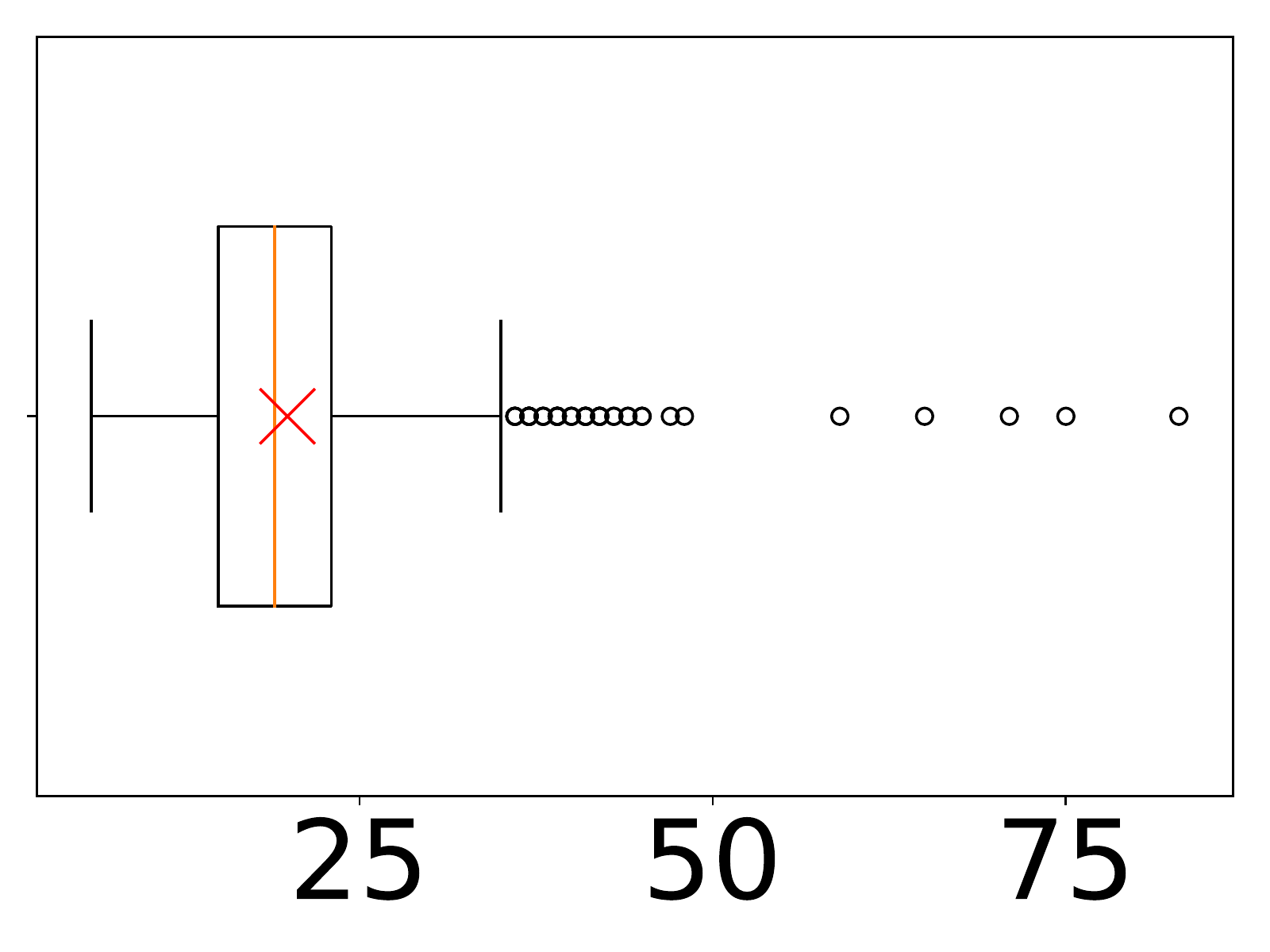}}
\subfigure[PROTEINS]{\includegraphics[width=.19\linewidth]{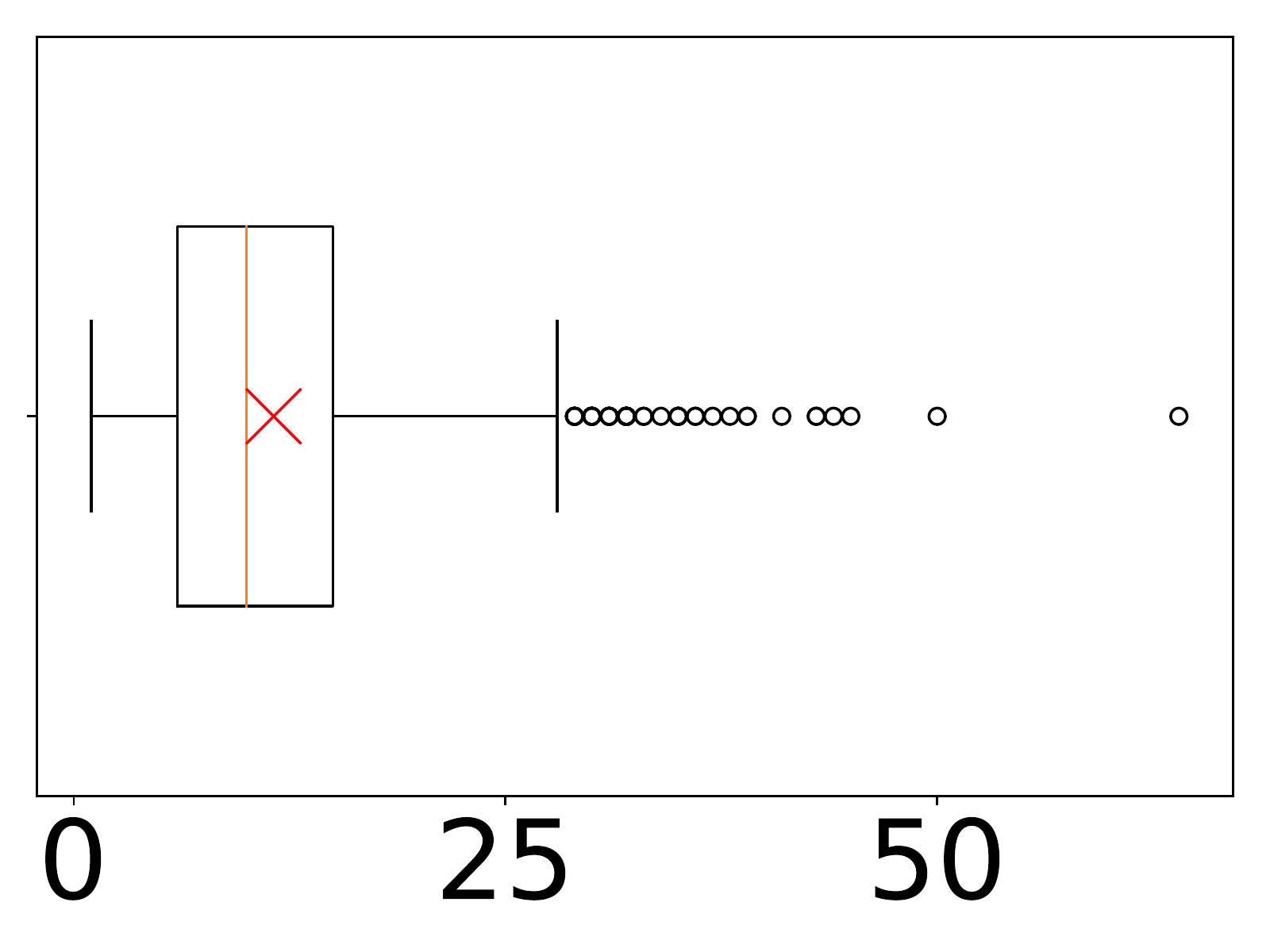}}
\subfigure[ENZYMES]{\includegraphics[width=.19\linewidth]{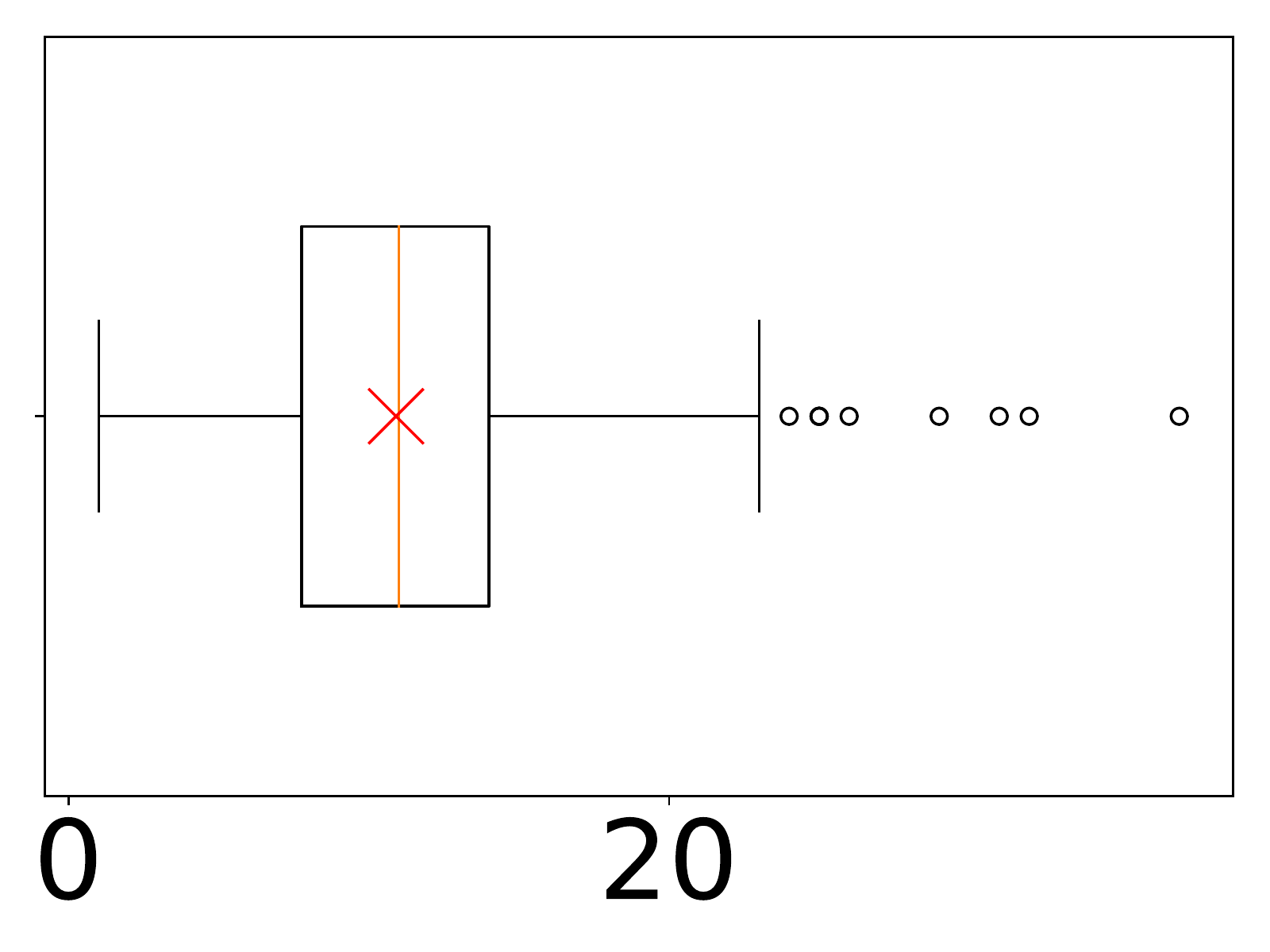}}
\subfigure[MUTAG]{\includegraphics[width=.19\linewidth]{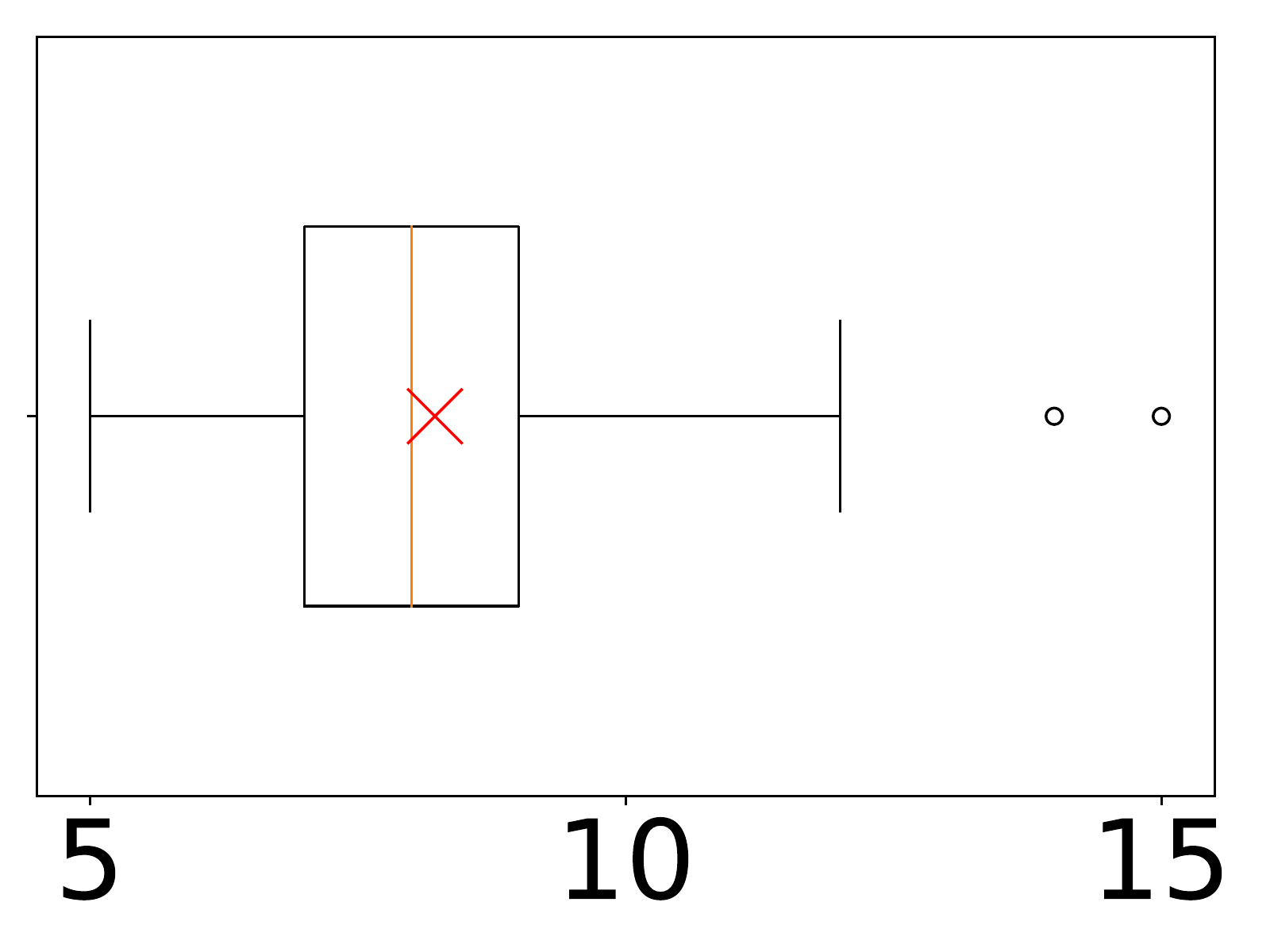}}
\caption{Box plot of the diameter distribution. The orange bar marks the median; the red cross the mean.}
\label{fig:stat2}
\end{figure*}

See Figures~\ref{fig:stat} and~\ref{fig:stat2}.

\end{document}